\documentclass{article}

\usepackage{microtype}
\usepackage{graphicx}
\usepackage{subcaption}
\usepackage{booktabs} 

\usepackage[draft]{hyperref}

\usepackage[utf8]{inputenc}
\usepackage{amsmath}
\usepackage{graphicx}
\usepackage{multirow}
\usepackage{wrapfig}
\usepackage{array}
\usepackage{amsfonts}
\usepackage{makecell}

\usepackage{xcolor}

\usepackage[accepted]{icml2019}

\icmltitlerunning{A Generative Model for Sampling High-Performance and Diverse Weights for Neural Networks}

\begin{document}

\twocolumn[
\icmltitle{A Generative Model for Sampling High-Performance and\\Diverse Weights for Neural Networks}



\icmlsetsymbol{equal}{*}

\begin{icmlauthorlist}
\icmlauthor{Lior Deutsch}{equal,ir}
\icmlauthor{Erik Nijkamp}{equal,uc}
\icmlauthor{Yu Yang}{uc}
\end{icmlauthorlist}

\icmlaffiliation{ir}{Independent Researcher}
\icmlaffiliation{uc}{Department of Statistics, UCLA}

\icmlcorrespondingauthor{Lior Deutsch}{sliorde@gmail.com}
\icmlcorrespondingauthor{Erik Nijkamp}{erik.nijkamp@gmail.com}
\icmlcorrespondingauthor{Yu Yang}{yy19970901@ucla.edu}

\icmlkeywords{Hypernetwork}

\vskip 0.3in
]



\printAffiliationsAndNotice{\icmlEqualContribution} 

\begin{abstract}
Recent work on mode connectivity in the loss landscape of deep neural networks has demonstrated that the locus of (sub-)optimal weight vectors lies on continuous paths. In this work, we train a neural network that serves as a hypernetwork, mapping a latent vector into high-performance (low-loss) weight vectors, generalizing recent findings of mode connectivity to higher dimensional manifolds. We formulate the training objective as a compromise between accuracy and diversity, where the diversity takes into account trivial symmetry transformations of the target network. We demonstrate how to reduce the number of parameters in the hypernetwork by parameter sharing. Once learned, the hypernetwork allows for a computationally efficient, ancestral sampling of neural network weights, which we recruit to form large ensembles. The improvement in classification accuracy obtained by this ensembling indicates that the generated manifold extends in dimensions other than directions implied by trivial symmetries. For computational efficiency, we distill an ensemble into a single classifier while retaining generalization.
\end{abstract}
 
\section{Introduction}
In recent years, generative methods such as Variational Autoencoders~(VAEs) \citep{kingma2013} and Generative Adversarial Networks~(GANs) \citep{NIPS2014_5423} have shown impressive results in generating samples from complex and high-dimensional distributions. The training of these generative models is data-driven, and an essential requirement is the existence of a rich enough data set, which faithfully represents the underlying probability distribution. A trained VAE decoder or GAN generator is a neural network which is operated by feeding it with an input (usually random), and it outputs an array of numbers, which represent, for example, pixel intensities of an image.

It is natural to suggest this idea be extended from images to neural networks, such that a trained generator would output numbers which represent weights for a neural network with a fixed target architecture and a fixed task, such as classifying a specific type of data. By ``weights'' we are referring to any trainable parameter of a neural network, including bias parameter. After all, just like images, a neural network is a structured array of numbers. Indeed, this was carried out in recent works \citep{krueger2017bayesian,louizos2017multiplicative}. The term \textit{hypernetwork} was coined \citep{ha2016hypernetworks,krueger2017bayesian} for a neural network that acts like such a generator. Applications include ensemble creation and Bayesian inference. Hypernetworks can also serve as a tool for a researcher to explore the loss function surface. 

A hypernetwork cannot be trained in a data-driven manner in the same sense as the aforementioned generative methods, since there is no rich data set of neural networks for each given target architecture and each task. Thus, there is no underlying probability distribution of neural networks. We, therefore, take a different approach, where we decide upon useful properties that such a distribution would have, and express them as loss function terms. The two useful properties are \textbf{accuracy} and \textbf{diversity}. The former implies that the generated neural networks achieve high accuracy in performing their intended tasks. The diversity property means that the hypernetwork could generate a big number of essentially different networks. Two networks are considered \textit{essentially different} if their weights differ by more than just trivial symmetry transformations.

Previous work \citep{krueger2017bayesian,louizos2017multiplicative} are set up in a Bayesian context, require a probabilistic interpretation of the target neural network's output, and harness Variational Inference (VI) \citep{hinton1993keeping,graves2011practical,kingma2017variational} to approximate the posterior of the weights given the data and sample from it. Our method improves upon some aspects of this approach. First, we do not require a probabilistic interpretation of the target network, which allows us to apply our method to a broader range of neural network types. Second, our loss function has an explicit hyperparameter that can be tuned to balance between accuracy and diversity. Lastly, our formalism allows different forms for the diversity loss term, while VI forces the use of entropy. Our formalism has VI as a private case, for specific choices of the hyperparameter and the diversity loss term.

\begin{figure}[t]
	\centering
	\begin{subfigure}{.5\textwidth}
		\centering
		\includegraphics[scale=.8]{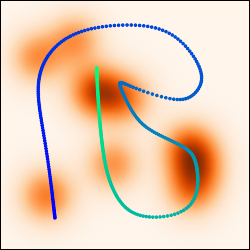}
		\caption{}
		\label{fig:toy_example_im}
	\end{subfigure}
	\begin{subfigure}{.5\textwidth}
		\centering
		\includegraphics[scale=1]{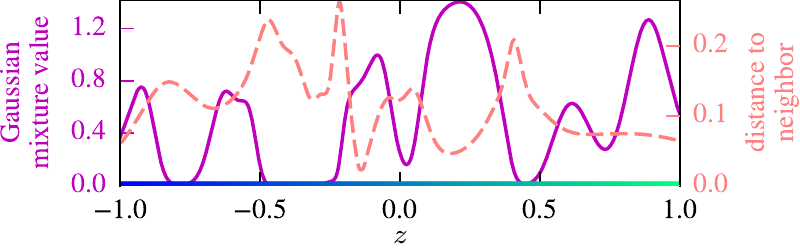}
		\caption{}
		\label{fig:toy_example_im}
	\end{subfigure}
	\caption{The hypernetwork on a toy problem in two dimensions. \textbf{(a)} Points generated by hypernetwork, where the background shades are the values of the Gaussian mixture. The curve (composed of points) is the hypernetwork's output for $400$ uniformly spaced points $z$ in the range $[-1,1]$. The points were colored so that they can be easily matched with their $z$ values in the other graphs in this figure. \textbf{(b)} The values of the Gaussian mixture (solid curve, left axis) along the path, and the distance of each point to its nearest neighbor (dashed curve, right axis).}
	\label{fig:toy_example}
\end{figure}

The main contributions of this work:
\begin{itemize}

\item We provide a simple template for hypernetwork loss functions, which have VI as a private case. The loss function is applicable to any type of neural network task - not only to networks with a probabilistic interpretation.

\item We show how to make the measure of the diversity of the generated networks more meaningful by taking symmetry transformations into account.

\item We describe a parameter sharing architecture which reduces the size of the hypernetwork.

\item We demonstrate for the first time (to our knowledge) a hypernetwork that can generate all the weights of a deep neural network, in such a way that all weights are statistically dependent. We show that the set of generated networks lies on a highly non-trivial manifold in weights space.

\item We distill an ensemble into a single neural network by stochastic approximation to improve computational efficiency while retaining the accuracy of the ensemble. 

\end{itemize}

\section{Related Work}\label{sec:related}
Recent papers have discussed the existence of connected one-dimensional regions in the space of weights of a neural network, with low loss values \citep{freeman2016topology,draxler2018essentially,garipov2018loss}.  \citep{freeman2016topology} show this for half-rectified networks with two layers. \citep{draxler2018essentially} and \citep{garipov2018loss} construct such paths using Nudged Elastic Bands and quadratic Bézier curves, respectively. It is natural to ask if we can capture higher dimensional manifolds as well.


Bayesian Hypernetworks (BHNs) \citep{krueger2017bayesian} and multiplicative normalizing flows with Gaussian posterior (MNFGs) \citep{louizos2017multiplicative} both transform a random input into the weights of a target neural network. These works formalize the problem in a Bayesian setting and use VI to get approximate samples from the posterior of the target network's weights. A key ingredient used by both is a normalizing flow (NF) \citep{rezende2015variational}, which serves as a flexible invertible function approximator. A NF requires that the input size is equal to the output size, and thus scales badly with the target networks' size. 

To reduce the number of parameters in the hypernetwork, BHNs and MNFGs employ reparameterizations of the weights. In BHN, the chosen reparameterization is weight normalization \citep{NIPS2016_6114}, and the NF produces samples only of the norms of the filters. The remaining degrees of freedom are trained to be constant (non-random). MNFG models the weights as a diagonal Gaussian when conditioned on the NF, with trainable means and variances. The NF acts as scaling factors on the means, one scale factor per filter (in the case of convolutional layers). The sources of randomness are thus the NF input and the per-weight Gaussian. The weights for different layers are generated independently. This limits the diversity of generated networks, since each layer needs to learn how to generate weights which give good accuracy without having any information about the other layers' weights. 

In contrast, in the current work, we don't use an NF, relying instead on MLPs and convolutions to create a flexible distribution. This makes it possible to use a small vector as the latent representation of the primary network, while generating all of its weights, without the need to use a restrictive model (e.g. Gaussian). The downside of this approach is that the MLPs are not invertible, which might cause the output manifold to have a lower dimension than the input, and which makes it necessary to use an approximation for the entropy of the output.

There is more prior art on using auxiliary neural networks for the task of obtaining the weights of a target network. Here we discussed the two most relevant to our work. In Appendix \ref{sec:related2} we review additional papers. Common to most of these papers is that there is no attempt for the generated weights to be diverse.

\section{Methods}\label{sec:methods}

\subsection{Hypernetworks}
Let $T\left(x;\theta\right):X \times \Theta \rightarrow Y$ be the neural network whose weights we want to generate, where $X$ is the input domain, $Y$ is the output domain and $\Theta$ is the set of trainable weight vectors of the network. We refer to $T$ as the \textit{target network architecture}, or more shortly as the \textit{target network}. Let $\mathcal{L}(\theta\vert p_{\text{data}})$ be the loss function associated with the target network, where $p_{\text{data}}\left(x,y\right)$, defined on $X \times Y$, is the data distribution. The standard practice for obtaining useful weights $\theta$ is to minimize $\mathcal{L}$ using backpropagation, where the gradients of $\mathcal{L}$ are estimated using batches of samples from the training data set. The outcome of this process is a single optimal vector of weights $\theta^*$. The set of possible outcomes may be very large, due to the prevalence of local minima and flat regions in the loss function surface \citep{choromanska2015loss,NIPS2014_5486} and due to symmetry transformations, such as permutation between filters and scaling of weights, which keep the network output unchanged.

The approach offered here is different. Instead of obtaining $\theta^*$ by directly minimizing $\mathcal{L}$, we obtain $\theta^*$ as the output of a \textit{hypernetwork}, which is a generator neural network $G\left(z;\varphi\right):Z \times \Phi \rightarrow \Theta$, where $Z$ is some input domain and $\Phi$ is the space of parameters (To reduce confusion, we refer to $\theta$ as ``weights'' and to $\varphi$ as ``parameters''; We occasionally use the notation $G\left(z\right)$ as a shorthand for $G\left(z;\varphi\right)$). We will draw the values for $z$ from a simple probability distribution $p_\text{noise}$. We train $G$ by minimizing a loss function $L(\varphi\vert p_{\text{noise}},p_{\text{data}})$ that depends on the combined network $T(x;G(z;\varphi))$.

\subsection{Loss Function}
\subsubsection{Two Components: Accuracy and Diversity}
To benefit from having a trained hypernetwork, it is not enough that $T\left(x;G\left(z;\varphi^*\right)\right)$ yields low values for $\mathcal{L}$. It is also important that the hypernetwork can generate essentially different networks (we define this in Section~\ref{sec:sym}). This forces us to let the hypernetwork $G(z)$ generate also sub-optimal weights. The reason is that $G(z)$ is a continuous function, and therefore if its image contains two distinct global optima, it must also contain a continuous path in weight space between them, and this path must also pass through sub-optimal weights (assuming there is no constant-loss path connecting the two optima) \citep{freeman2016topology}. Thus, we train $\varphi^*$ by minimizing over a loss function which includes two terms:
\begin{eqnarray} \label{eq:l}
 L\left( \varphi\vert p_\text{noise},p_\text{data} \right) &=& \lambda\nonumber L_\text{accuracy}\left( \varphi \vert p_\text{noise},p_\text{data}\right)\\
 &+&  L_\text{diversity}\left( \varphi\vert p_\text{noise}\right),
\end{eqnarray}
where $L_\text{accuracy}$ depends on $\mathcal{L}$. An obvious choice is 
\begin{equation*} \label{eq:l_acc}
L_\text{accuracy}\left( \varphi \vert p_\text{noise},p_\text{data}\right)=\mathbb{E}_{z \sim p_\text{noise}} \mathcal{L}\left( G\left(z;\varphi\right)\vert p_\text{data} \right). 
 \end{equation*}
$L_\text{diversity}$ should ensure that there is high diversity in the results of $G$ as a function of $z$, obviating the risk of undergoing ``mode collapse''. $\lambda>0$ is a hyperparameter that balances the two losses (Strictly speaking, $\lambda$ should not be called a hyperparameter, since it is not a variable such as the number of layers or a regularization coefficient which should be tuned so as to minimize the validation loss. $\lambda$ is an essential part of the validation loss function itself. However, we do not make this distinction in the following).

For the diversity term, we will use (the negative of) the entropy of the generated weights. Other possibilities include the variance, or diversity terms that are used in texture synthesis \citep{li2017diversified} and feature visualization \citep{olah2017feature}. During training, we estimate the entropy of a minibatch of generated weights using a modified form of the Kozachenko-Leonenko estimator \citep{kozachenko1987sample,kraskov2004estimating} (see Appendix \ref{sec:entropy}).

The loss function in equation (\ref{eq:l}) is heuristically simple to justify. However, it is illuminating to see how it can also be obtained in another way, as the relaxation of an optimistic choice for the distribution of $G(z)$. This is described in Appendix \ref{sec:relaxation}.

\begin{figure}[t]
	\centering
	\begin{subfigure}{.45\linewidth}\centering\includegraphics[scale=1]{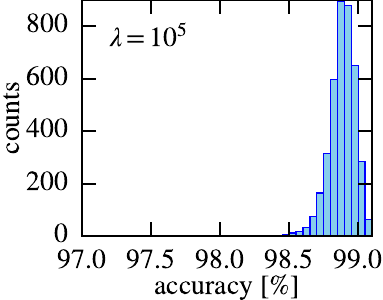}\caption{}\label{fig:histogram_high}\end{subfigure}
	\begin{subfigure}{.45\linewidth}\centering\includegraphics[scale=1]{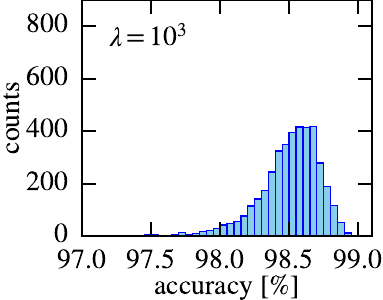}\caption{}\label{fig:histogram}\end{subfigure}
	\begin{subfigure}{\linewidth}\centering\includegraphics[scale=1]{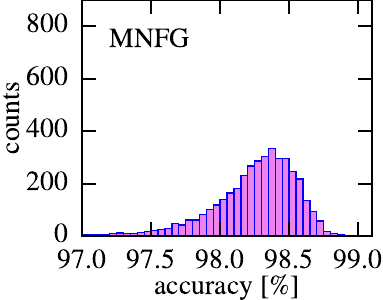}\caption{}\label{fig:histogram_mnf}\end{subfigure}
	\caption{Histograms of accuracies of generated networks. \textbf{(a)} Our hypernetwork, $\lambda=10^5$. \textbf{(b)} Our hypernetwork, $\lambda=10^3$. \textbf{(c)}~MNFG.}
	\label{fig:histograms}
\end{figure}

\subsubsection{Taking Symmetries Into Account}\label{sec:sym}
An increase in entropy may not always translate to an increase in diversity, since the diversity that we are interested in is that of \textit{essentially different} outputs. Two weight vectors $\theta_1,\theta_2\in \Theta$ are considered \textit{essentially different} if there is no trivial symmetry transformation that transforms $\theta_1$ into a close proximity of $\theta_2$, where the proximity is measured by some metric on $\Theta$. Symmetry transformations are functions $S:\Theta \rightarrow \Theta$ such that $T\left(x;\theta\right)=T\left(x;S(\theta)\right)$ for all $x\in X$ and all $\theta \in \Theta$. The trivial symmetry transformations include any composition of the following:

\begin{itemize}
\item \textbf{Scaling} - If the target network is a feed-forward convolutional network with piecewise-linear activations such as ReLU \citep{jarrett2009best,glorot2011deep} or leaky-ReLU \citep{maas2013rectifier}, then scaling the weights of one filter by a positive factor, while unscaling the weights of the corresponding channel in all filters in the next layer by the same factor, keeps the output of the network unchanged.

\item \textbf{Logits' bias} - If a network produces logit values which are fed into a softmax layer, then adding the same number to all logit values does not change the values of the softmax probabilities.

\item \textbf{Permutation} - Permuting the filters in a layer, while performing the same permutation on the channels of the filters in the next layer.
\end{itemize}

A hypernetwork that generates weight vectors that differ only by a trivial symmetry transformation should not score high on diversity, even though it may have high entropy. To deal with this problem, we use \textit{gauge fixing} (a term borrowed from theoretical physics), which breaks the symmetry by choosing only one representative from each equivalence class of trivial symmetry transformations. This can be realized by a function $\mathcal{G}:\Theta \rightarrow \Theta$ which transforms any weight vector to its equivalence class representative: $\mathcal{G}(\theta_1)=\mathcal{G}(\theta_2)$ if and only if $\theta_1$ and $\theta_2$ are related by a trivial symmetry transformation. Therefore, we use the following form for the entropy in the diversity term:
\begin{equation} \label{eq:l_div}
L_\text{diversity}\left(\varphi\vert p_\text{noise}\right) = -\mathbb{H}_{z \sim p_\text{noise}}\left[\mathcal{G}\left(G\left(z;\varphi\right)\right)\right],
\end{equation}
where $\mathbb{H}$ is the entropy of its argument. We choose $\mathcal{G}$ to break the symmetries in scaling and in logits' bias. The former is broken by requiring ${\sum_k\left(\theta_{l,i}[k]\right)^2=n_{l,i}}$, where $\theta_{l,i}[k]$ is the $k$'th element of the $i$'th filter of the $l$'th layer, and $n_{l,i}$ is the number of elements in the filter, including the bias term. The sum is over all elements of the filter. This constraint is applied to the layers $1\leq l \leq m-1$ where $m$ is the number of layers, and for all filters $i$ in the layer. It is imposed on all but the last layer, since this is the freedom we have under the scaling symmetry transformation. To see this, we can take an arbitrary network, and repeat the following process, starting with $l=1$ and then incrementing $l$ by $1$: we scale the filters of the $l$'th layer to obey the constraint, and unscale the filters of the $l+1$ layer accordingly to keep the output unchanged. This process cannot be performed when $l=m$, since for the last layer there is no next layer to do the unscaling on. The logits' bias symmetry is broken by requiring $\sum_i\theta_{m,i}[\text{bias}]=0$, where the summand  is the bias term of the filter. Incorporating permutation symmetry is also possible, for example by lexicographically sorting filters in a layer. However, we decide to ignore this symmetry, due to implementation constraints. Note that the permutations form a discrete group, and it is harder for a continuous generator to fail by generating discrete transformations.

\begin{figure}[t]
	\centering
	\begin{subfigure}{.2\textwidth}
		\centering
		\includegraphics[scale=0.5]{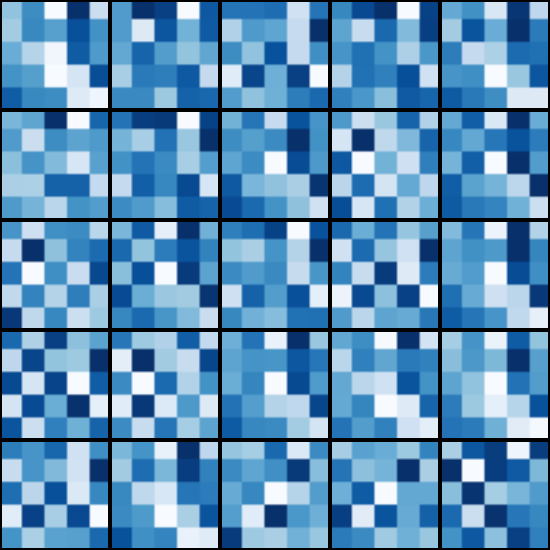}
		\caption{}
		\label{fig:visual1}
	\end{subfigure}
	\begin{subfigure}{.2\textwidth}
		\centering
		\includegraphics[scale=0.5]{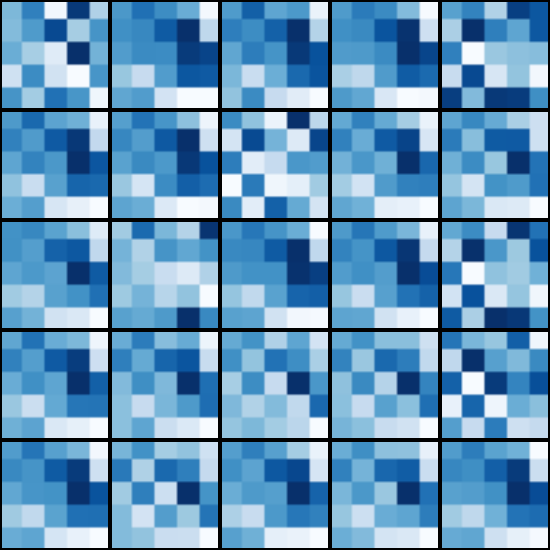}
		\caption{}
		\label{fig:visual2}
	\end{subfigure}
	\begin{subfigure}{.2\textwidth}
		\centering
		\includegraphics[scale=0.5]{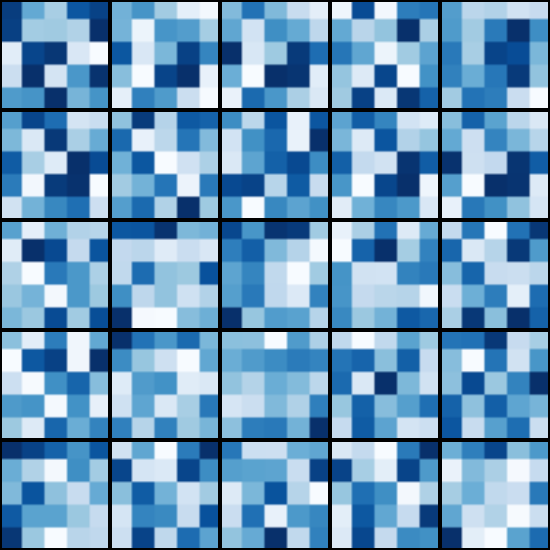}
		\caption{}
		\label{fig:visual3}
	\end{subfigure}
	\begin{subfigure}{.2\textwidth}
		\centering
		\includegraphics[scale=0.5]{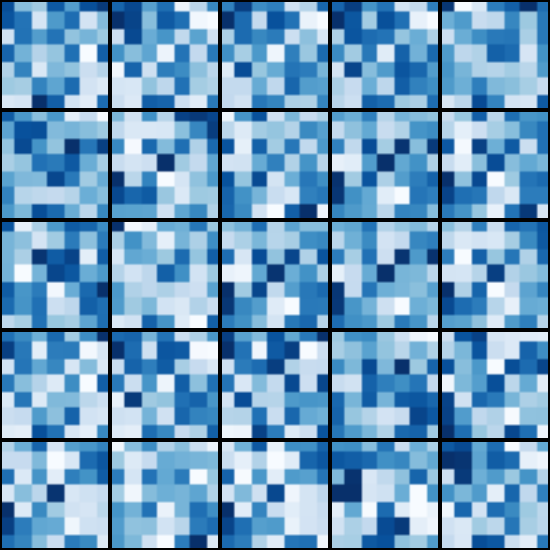}
		\caption{}
		\label{fig:visual4}
	\end{subfigure}
	\caption{Examples of samples of filter slices. Each of the four figures contains $25$ samples for one specific filter slice. \textbf{(a)} first layer. \textbf{(b)} first layer (different filter). \textbf{(c)} second layer. \textbf{(d)} third layer.}
	\label{fig:visual}
\end{figure}

\subsubsection{Diversity Measures Which are not Affected by Symmetries}\label{sec:sym}
An alternative approach for the diversity loss term is to measure diversity in the \textit{results} of the generated networks, as opposed to diversity in their weights (this means that $L_\text{diversity}$ will depend on $p_\text{data}$ and well as $p_\text{noise}$). This approach will inherently assign low diversity when symmetries are generated.  Here are two ways to do this:

\begin{itemize}
	\item Instead of calculating the entropy of the weight vector, calculate the entropy of the concatenation of the outputs of the target network for various inputs. This can be written informally as
	\begin{equation} \label{eq:entropy_sm}
		\mathbb{H}_{z \sim p_\text{noise}}\left[\text{concat}_{x \sim p_\text{data}} T(x;G(z;\varphi))\right].
	\end{equation}
	\item We expect high diversity to yield an increase in prediction accuracy for an ensemble of generated networks, compared to the individual networks. This can be translated into a loss function by aggregating the results of different generated networks (e.g by averaging), and applying the target network's loss criteria on this aggregation.
\end{itemize}

\subsubsection{The Case of Classification, and the Relation to Variational Inference}
In Section~\ref{sec:experiments}, we describe experiments where the target network performs classification. In this case, the outputs of the target network are probability distributions over the finite set of classes. Thus, $T\left(x;\theta\right)_i$ is the probability for the $i$'th class for input $x$. We can rewrite this as $p(i\vert x; \theta)$. Typically, the loss function is taken to be the negative mean log likelihood:
\begin{equation} \label{eq:l_class}
\mathcal{L}(\theta\vert p_{\text{data}}) = -\mathbb{E}_{(x,y) \sim p_\text{data}} \log p(y\vert x; \theta) ,
\end{equation}	
where for simplicity we assumed here that the data set contains only deterministic distributions $y$, i.e. $y$ is a one-hot encoding of a class, and we can identify between $y$ and its class.

Combining equations (\ref{eq:l} - \ref{eq:l_class}) while ignoring the gauge fixing function and setting $\lambda=n$, where $n$ is the size of the data set, we obtain:
\begin{equation}  \label{eq:ll}
\begin{split}
 L\left( \varphi\vert p_\text{noise},p_\text{data} \right) & = \\
 & -\mathbb{E}_{z \sim p_\text{noise},(x,y) \sim p_\text{data}} n\log p(y\vert x; G\left(z;\varphi\right)) \\
 & -\mathbb{H}_{z \sim p_\text{noise}}\left[G\left(z;\varphi\right)\right].
\end{split} 
\end{equation} 
This is very similar to the VI objective \citep{krueger2017bayesian,louizos2017multiplicative}, as we describe in Appendix \ref{sec:vi}. However, equation (\ref{eq:l}) is more general in that it does not require the outputs of the target network to be probability distributions, it allows us to use diversity terms other than entropy, which could include gauge fixing, and it incorporates the hyperparameter $\lambda$ that enables us to control the balance between accuracy and diversity (However, we believe that the appearance of $\lambda$ is consistent with VI, as we explain in Appendix \ref{sec:vi2}).

\subsection{Architecture}

\begin{figure}
	\centering
	\includegraphics[scale=1.08]{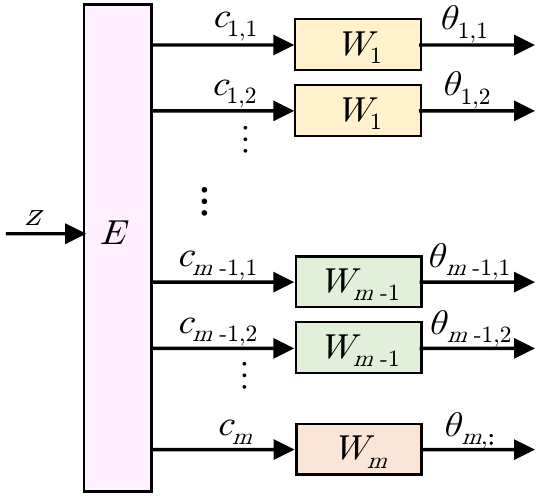}
	\caption{Architecture block diagram of the hypernetwork with latent variable $z$, extractor $E$, codes $c$, weight-generators $W$, and generated weights $\theta$.}
	\label{fig:architecture}
\end{figure}

We assume that the target network $T$ is convolutional with $m$ layers, where within each layer the filters have the same size. $\theta_{l,i}$ are the weights for the $i$'th filter of the $l$'th layer. If the hypernetwork $G$ were a fully connected multilayer perceptron (MLP), it would scale badly with the dimension of the weights of $T$. We therefore utilize parameter sharing in $G$ by giving it a convolutional structure, see Fig.~\ref{fig:architecture}.

The input $z$ for the hypernetwork is drawn from a prior distribution such as Uniform or Gaussian and is fed into a fully-connected sub-network $E$, which we call the \textit{extractor}, whose output is a set of \textit{codes} $c_{l,i}$, where $l=1,..,m$. The code $c_{l,i}$ is a latent representation of $\theta_{l,i}$. The code $c_{l,i}$ is then fed into the \textit{weight generator} $W_l$, which is another fully connected network which generates the weights $\theta_{l,i}$ for the target network. We emphasize that the same weight generator is re-used for all filters in a certain layer of $T$, thus reducing the overall number of parameters in the hypernetwork. We consider the weights at the input of a fully connected neuron as a filter, whose receptive field is the entire previous layer.

\begin{figure}[t]
	\centering
	\begin{subfigure}{.2\textwidth}
		\centering
		\includegraphics[scale=0.5]{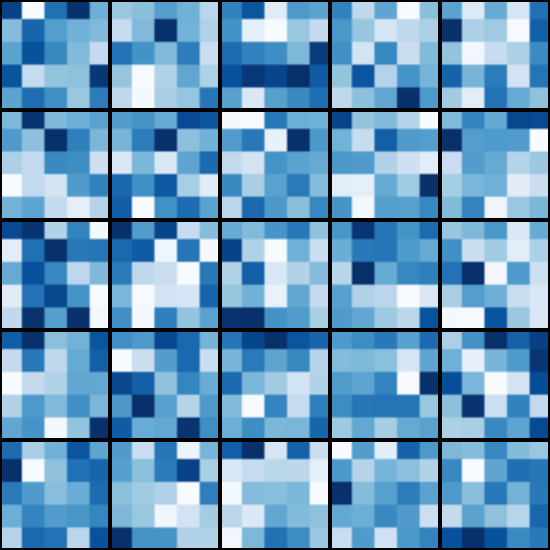}
		\caption{}
		\label{fig:visual1_mnf}
	\end{subfigure}
	\begin{subfigure}{.2\textwidth}
		\centering
		\includegraphics[scale=0.5]{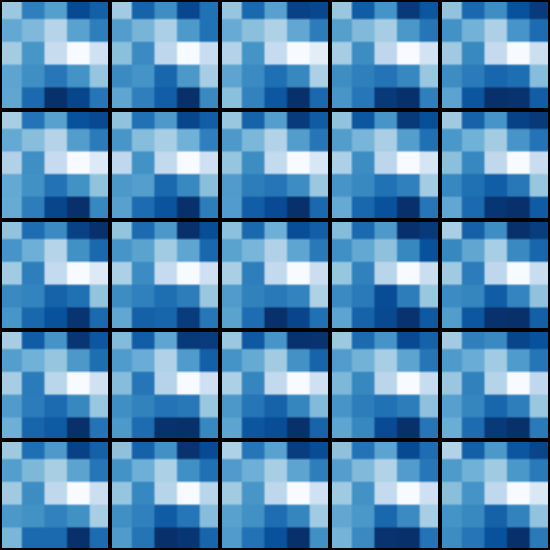}
		\caption{}
		\label{fig:visual2_mnf}
	\end{subfigure}
	\begin{subfigure}{.2\textwidth}
		\centering
		\includegraphics[scale=0.5]{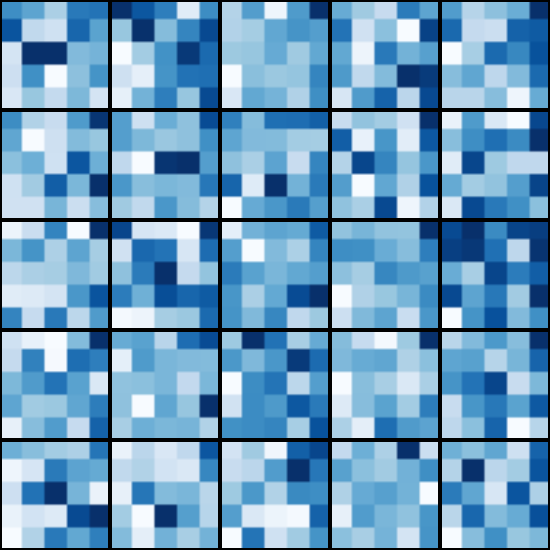}
		\caption{}
		\label{fig:visual3_mnf}
	\end{subfigure}
	\caption{Examples of samples of filter slices \textbf{for MNFG}. Each of the three figures contains $25$ samples for one specific filter slice. \textbf{(a)} first layer. \textbf{(b)} first layer (different filter). \textbf{(c)} second layer.}
	\label{fig:visual_mnf}
\end{figure}

$W_l$ can be seen as a convolutional non-linear filter with a receptive field that is a single code for the layer $l$. To allow flexible use of high-level features, we treat all weights in the last layer of $T$ as one ``filter'', thus using $W_m$ only once.

To ease the optimization of the hypernetwork $G$, we may introduce residual connections and batch-normalization in extractor network $E$ and weight generators $W_l$.

\section{Experiments}\label{sec:experiments}
All code used to run the experiments can be found online\footnote{\url{https://github.com/sliorde/generating-neural-networks-with-neural-networks}}. Please refer to the code for a full specification of hyperparameter values and implementation details.

\subsection{Toy Problem}
We start with a toy problem which is easy to visualize. Instead of generating weights for a neural network, we generate a two-dimensional vector. The goal is for the generated vectors have high values on a specified Gaussian mixture, and also to obtain high diversity. We take the input $z$ to the hypernetwork to have dimension $1$, to demonstrate the case where the hypernetworks output manifold has a lower dimensionality than the weight space. This problem does not have symmetries, so we do not include gauge fixing in the loss. Instead, we use a conventional $\ell_2$ regularization loss. The hypernetwork is taken as an MLP whose hidden layers have sizes $30$, $10$ and $10$. The final distribution learned by the hypernetwork is displayed in Fig.~\ref{fig:toy_example}. We see that the one-dimensional distribution is supported on a path which passes through all peaks of the Gaussian mixture. Inevitably, the path must pass through regions with low values of the Gaussian mixture. However, in these regions, the density is lower.

\begin{figure}[t]
	\centering
	\includegraphics[width=1.01\linewidth]{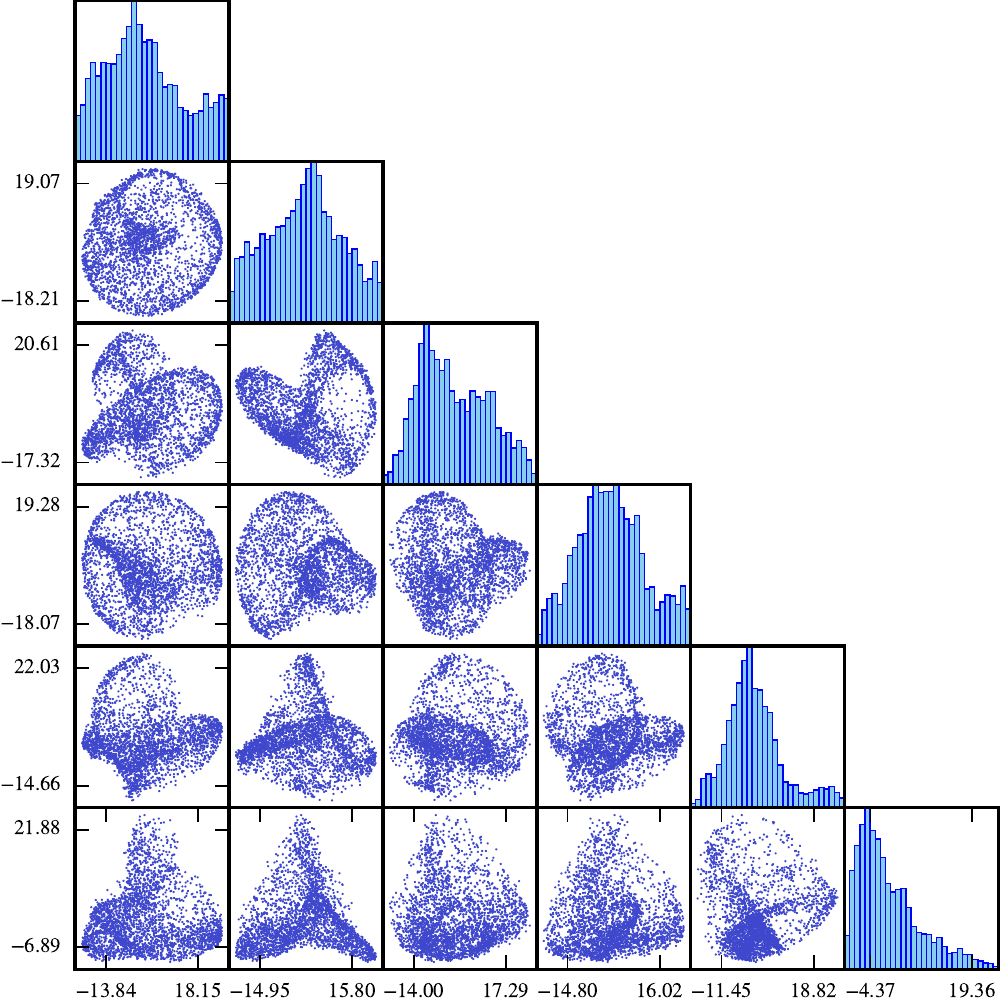}
	\caption{Scatter plots of the generated weights for a specific second layer filter, in PCA space. The $(i,j)$ scatter plot has principal component $i$ against principal component $j$.}
	\label{fig:mnist_layer2_filt_pca}
\end{figure}

\begin{figure}[t]
	\centering
	\begin{subfigure}{\linewidth}\centering\includegraphics[scale=1.1]{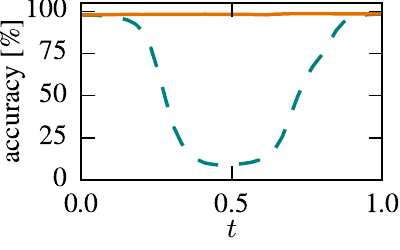}\end{subfigure}
	\begin{subfigure}{\linewidth}\centering\includegraphics[scale=1.1]{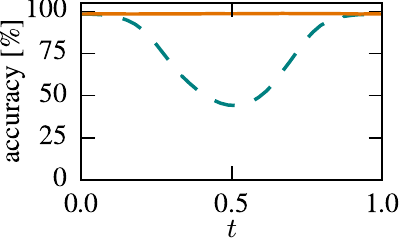}\end{subfigure}
	\begin{subfigure}{\linewidth}\centering\includegraphics[scale=1.1]{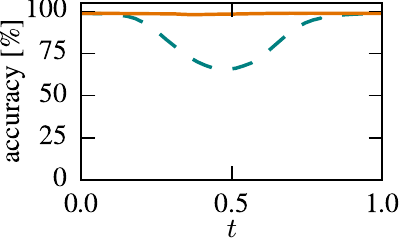}\end{subfigure}
	\caption{Accuracies along three paths, with end points $z_1$, $z_2$ sampled at random. The dashed lines are the direct paths, and the solid lines are the interpolated paths.}
	\label{fig:paths}
\end{figure}

\subsection{MNIST}
We take the target network architecture $T$ to be a simple four layer convolutional network for classifying images of the MNIST data set \citep{lecun1998gradient}. The full target network specification is displayed in Appendix \ref{sec:mnist_target}. The total number of weights in the network is $20,018$. This network can easily be trained to achieve an accuracy of over $99\%$ on the validation set.

For the hypernetwork, we take the input vector $z$ to be $300$ dimensional, drawn from a uniform distribution. The extractor and weight generators have three layers each. The total number of parameters is $633,640$. For a detailed specification, including training details, see Appendix \ref{sec:mnist_hyper}.

We compare our results to MNFG~\citep{louizos2017multiplicative}. We used code that is available online\footnote{\url{https://github.com/AMLab-Amsterdam/MNF_VBNN}}, and modified it so that it generates the same target networks as our hypernetwork.

\subsubsection{Accuracy}
The validation set accuracy of generated networks depends on the hyperparameter $\lambda$. Figs. \ref{fig:histogram_high} and \ref{fig:histogram} display histograms of the accuracies of the generated weights, for $\lambda=10^5$ and $\lambda=10^3$ respectively. We also show the corresponding histogram for MNFG in Fig. \ref{fig:histogram_mnf}. From now on, we will use $\lambda=10^3$, which yields lower accuracies, but they are more comparable to the results of MNFG.

\begin{table*}[t]
	{\renewcommand{\arraystretch}{1.4}%
	\begin{center}
		\begin{tabular}{ | c | c | c c c | c c c c | }
			\hline
			\textbf{Dataset} & \textbf{Model} & \multicolumn{3}{c|}{\textbf{SGD}} &
			\multicolumn{4}{c|}{\textbf{Hypernetwork}} \\
			& & (a) 1 Sample & \multicolumn{2}{c|}{(b) 10 Samples} & \multicolumn{2}{c}{(c) w/o Entropy} & \multicolumn{2}{c|}{(d) w/ Entropy} \\
			& & & Mean & Majority & Mean & Majority & Mean & Majority \\
			\specialrule{.1em}{.05em}{.05em} 
			\multirow{2}{*}{CIFAR-10} & LeNet & 77.36\% & 74.95\% & 80.60\% & 76.26\% & 76.47\% & 72.34\% & 78.21\% \\  
			& ResNet-32 & 93.57\% & 91.03\% & 94.96\% & 88.20\% & 88.26\% & 83.61\% & 90.37\% \\  
			\hline
		\end{tabular}
	\end{center}
	}
	\caption{Test accuracies on the CIFAR-10 \iffalse and CIFAR-100\fi dataset for \textbf{(a)} SGD with decaying learning rate as reference and ensembles where \textbf{(b)} samples are drawn from a SGD dynamics, and, \textbf{(c-d)} samples drawn by ancestral sampling from the hypernetwork.}
	\label{table:cifar}
\end{table*}

\subsubsection{Diversity}
We explore the diversity in a few different ways. The histograms in Fig. \ref{fig:histograms} give an initial indication of diversity by showing that there is variance in the generated networks' accuracies.

\paragraph{Visual Inspection.} In Fig.~\ref{fig:visual} we show images of different samples of the generated filters. By visual inspection we see that different samples can result in different forms of filters. However, it is noticeable that there are some repeating patterns between samples. For comparison, we show corresponding images for MNFG in Fig.~\ref{fig:visual_mnf}, where we gauged the filters generated by MNFG just as our own. We see that MNFG yields high diversity for most filters (e.g. Fig.~\ref{fig:visual1_mnf}), but very low diversity for others (Fig. \ref{fig:visual2_mnf}), mainly in the first layer. We see this phenomena also in our generated filters only to a lesser extent. We hypothesize that a possible mode of failure for a hypernetwork is when it concentrates much of its diversity in specific filters, while making sure that these filters get very small weighs in the next layer, thereby effectively canceling these filters. Future work should consider this mode of failure in the diversity term of the loss function.

\paragraph{Scatter.} One may wonder whether our method of weight generation is equivalent to trivially sampling from $\mathcal{N}(\theta_0,\Sigma)$, for some optimal weight vector $\theta_0$ and constant $\Sigma$. To see that this is not the case, we view the scattering of the weight vectors using principle component analysis (PCA). This is shown in Fig.~\ref{fig:mnist_layer1_filt_pca} and in Appendix \ref{sec:pca}. We see that the hypernetwork learned to generate a distribution of weights on a non-trivial manifold, with prominent one-dimensional structures. Scatter graphs for MNFG are displayed in Appendix \ref{sec:pca_mnf}.

The construction of connected regions in weight space, with low accuracy loss values, was recently discussed in \citep{freeman2016topology,draxler2018essentially,garipov2018loss} for the case of one-dimensional regions. Here we see that this can be done also for higher dimensional manifolds.

\paragraph{Paths in Weight Space.} For two given input vectors $z_1$ and $z_2$, we define two paths which coincide in their endpoints: The direct path $\{G(z_1)t+G(z_2)(1-t) \mid t\in[0,1] \}$, and the interpolated path $\{G(z_1t+z_2(1-t)) \mid t\in[0,1] \}$. We expect high accuracy along the interpolated path. Whether the direct path has high accuracy depends on the nature of the generated manifold. For example, if the diversity is achieved only via random isotropic noise around a specific weight vector, the direct path would have high accuracy. In Fig. \ref{fig:toy_example} we see that this is not the case. The analogous graph for MNFG, displayed in Appendix~\ref{sec:path_mnf}, shows that for MNFG the direct path does give high accuracy.

\paragraph{Ensembles.} We compare the accuracy of the generated networks with the accuracy of ensembles of generated networks. If the generated classifiers are sufficiently different, then combining them should yield a classifier with reduced variance~\citep{friedman2001elements}, and therefore lower error. We created $20$ ensembles, each of size $200$, and we take their majority vote as a classification rule. The result is that the average accuracy of the ensembles on the validation set was $99.14\% $, which is higher than typical results that we see in the histogram in Fig.~\ref{fig:histograms}. For MNFG, the same experiment gives an average ensemble accuracy of $99.28\%$.

\paragraph{Adversarial Examples.} We show that using ensembles of generated networks can help reduce the sensitivity to adversarial examples \citep{szegedy2013intriguing,goodfellow2014explaining}. The experiment was conducted by following these steps:  \textbf{1)} Use the hypernetwork to generate a weight vector $\theta$. \textbf{2)} Sample a pair $(x,y)$ of image and label from the validation set. \textbf{3)} Randomly pick a new label $y^\prime \neq y$ to be the target class of the adversarial example.\textbf{4)} Use the fast gradient method\citep{goodfellow2014explaining} to generate adversarial examples. Do this for perturbation sizes $\epsilon$ in the range $0$ to $0.24$, as fractions of the dynamic range of an image ($8$ bits of grayscale). \textbf{5)} Test which values of $\epsilon$ yield adversarial examples that fool the classifier with weights $\theta$. \textbf{6)} Use the hypernetwork to generate an ensemble of $100$ classifiers, and test for which values $\epsilon$ the adversarial examples created in step 4 fool the ensemble. We repeat this experiment over all images in the validation set. The results are shown in Fig. 000, together with results for MNFG. We see that the probability of success for an adversarial attack is reduced when using ensembles.

\begin{figure}[t]
	\centering
	\begin{subfigure}{\linewidth}
		\centering
		\includegraphics{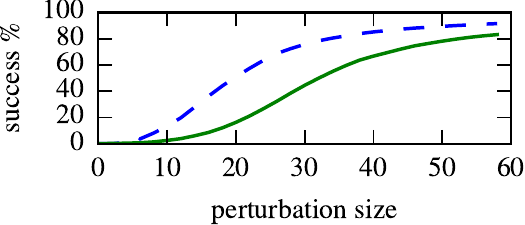}
		\caption{}
		\label{fig:adversarial}
	\end{subfigure}
	\begin{subfigure}{\linewidth}
		\centering
		\includegraphics{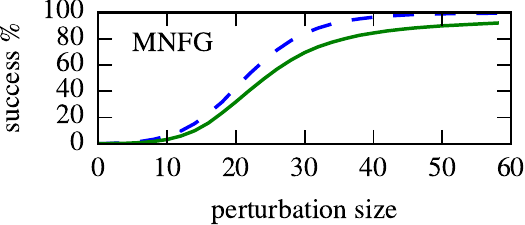}
		\caption{}
		\label{fig:adversarial_mnf}
	\end{subfigure}
	\caption{The success probability of adversarial examples created with the fast gradient method with a given perturbation size, against an ensemble of $100$ generated networks. The dashed lines are the single classifier, and the solid lines are ensembles. \textbf{(a)} our hypernetwork. \textbf{(b)} MNFG. }
	\label{fig:adversarials}
\end{figure}

\subsection{CIFAR-10}
We empirically validate the diversity of the generated target networks by comparing various ensembling methods and the scalability of the hypernetwork by introducing skip connections. As such, we train the hypernetwork with LeNet ($69,656$ parameters) and ResNet-32 ($466,906$ parameters) on the CIFAR-10~\citep{krizhevsky2014cifar} dataset. We refer Appendix~\ref{sec:cifar_hyper} for the specifications of the hypernetworks.

\subsubsection{Ensembles}
Table~\ref{table:cifar} depicts the maximum test accuracy for target networks trained by Stochastic Gradient Descent (SGD) and our hypernetwork. That is, (a) we train both models by SGD with decaying learning-rate to achieve close-to state-of-the-art-performance, (b) form an ensemble of 10 models by treating the SGD with constant learning-rate as a sampler \citep{mandt2017stochastic} from the weight posterior and report mean and majority vote accuracy, (c)  an ensemble of 10 models sampled from the hypernetwork trained without entropy term, (d) and with entropy term~(\ref{eq:entropy_sm}) as ablation study.

\paragraph{Ablation.}The result indicates that the hypernetwork can learn diverse weights and scale to larger target networks. For SGD ensembles of independent chains, as expected, the diversity in the weights leads to consistently highest test accuracy. For LeNets, the hypernetwork appears to learn a patch of the weight manifold when including the entropy term as objective and a sampled ensemble outperforms the accuracy of SGD on a single model by 0.85 on CIFAR-10. Including the entropy term decreases the mean accuracy while increasing the ensemble accuracy. For ResNet-32, the entropy term increases diversity in the weights, but the ensemble does not achieve state-of-the-art performance. We conjecture with further fine-tuning the performance on ResNet-32 can be improved.
 


\subsection{Distillation}
It is desirable to distill an ensemble of target networks into a single classifier for computational efficiency while retaining generalization. Polyak-Ruppert averaging~\citep{polyak1990new, ruppert1988efficient} is in the form of averaging iterates $\{\theta_{t}\}$ of a dynamics such that $\bar{\theta}_t = \frac{1}{t} \sum_{i=0}^{t-1} \theta_i$. We propose to substitute iterates of the trajectory by ancestral samples $\{\theta_i=G(z_i), z_i\sim\mathcal{N}(0,\sigma I)\}_{i=1}^n$. For LeNet on CIFAR-100, the ensemble accuracy of 45.42\% drops to 45.41\% under Polyak averaging. For ResNet-32 on CIFAR-10, the accuracy drops from 90.02\% to 89.46\%.

\section{Discussion}\label{sec:discussion}
In this work, we have shown how a hypernetwork can be trained to generate accurate and diverse weight vectors. Important directions of further inquiry are: Is there a certain gauge which yields better training? What are the performances of other diversity terms? How should diversity be evaluated? What are good methods for initializing parameters for the hypernetwork?

In answering the latter question, one should notice that popular methods (for example, \citep{glorot2010understanding,he2015delving}), use the $\text{fan}_\text{in}$ and $\text{fan}_\text{out}$ of a unit, but in the case of a weight generator sub-network in our  architecture, we may want to take into account also the $\text{fan}_\text{in}$ and $\text{fan}_\text{out}$ of the generated filter.

It is critical to find architectures for hypernetworks which scale better with the size of the target network. Ideally, hypernetworks will have fewer parameters than their target networks, and therefore could be used as a compressed version of the target network.

In the future, applications of the proposed hypernetwork like transfer learning or generating sparse weight vectors for target networks should be explored.

\bibliography{references}
\bibliographystyle{icml2019}

\clearpage
\appendix
\section{Additional Related Work}\label{sec:related2}
There are many methods of using auxiliary neural networks for the task of obtaining the weights of a target network. An early example is fast weights \citep{schmidhuber2008learning}, where the target network is trained together with an auxiliary memory controller(MC) network, whose goal is to drive changes in the weights of the target network. Although the overall architecture of this approach is quite similar to hypernetworks as presented here, the approach differs in two important aspects: \textbf{(a)} Fast weights are designed specifically as alternatives to recurrent networks for temporal sequence processing, and therefore the MC can never be decoupled from the target network, even after the MC has done a computation; \textbf{(b)} The input to the MC is just the input to the target network, and there is no attempt to generate diverse iid samples of the target network weights.

Under the paradigm of \textit{learning to learn}, the works \citep{andrychowicz2016learning,li2016learning} have trained auxiliary neural networks to act as optimizers of a target network. The optimizers apply all the updates to the target network's weights during its training. As opposed to the method presented here, these trained optimizers can generate weights to the target network only by receiving a long sequence of batches of training examples. On the other hand, these optimizers can generalize to various loss functions and problem instances.

In \citep{bertinetto2016learning}, the auxiliary network is used for one-shot learning: It receives as input a single training example, and produces weights for the target network. This is different from hypernetworks, where the input is a latent representation of the target network, and the generated weights are not seen as a generalization from a single training example. A related method is dynamic filter networks \citep{de2016dynamic}, where the auxiliary network is fed with the same input as the target network, or with a related input (such as previous frames of a video). The goal is to make the target network more adaptive to the instantaneous input or task, rather than to generate diverse versions of the target network which are on an equal footing.

It has been shown \citep{NIPS2013_5025} that for common machine learning tasks, there is a redundancy in the raw representation of the weights of several neural network models. \citep{ha2016hypernetworks} exploits this fact to train a small neural network that generates the weights for a larger target network. This has the advantage of reduced storage size in memory, and can also be seen as a means of regularization. However, the weight generating network in \citep{ha2016hypernetworks} does not have a controllable input, and therefore it cannot be used to generate diverse random samples of weights.

In \citep{lorraine2018stochastic}, a hypernetwork is presented whose input are hyperparameter values, and it generates weights for the target network, which correspond to the hyperparameter.

In HyperNEAT \citep{stanley2009hypercube} an auxiliary neural network is evolved using a genetic algorithm. This auxiliary network encodes the weights of the target network in the following way: The neurons of the target network are assigned  coordinates on a grid. The weight between every pair of neurons is given by the output of the auxiliary network whose input are the coordinates of the two neurons. Generating the weights for the entire target network requires reapplication of the auxiliary network to all pairs of neurons. The goal of HyperNEAT is to generate networks with large scale that exhibit connectivity patterns which can be described as functions on low dimensions. There is no emphasis on generating a diverse set of such networks.

In \citep{zoph2016neural}, the auxiliary neural network is a recurrent network, trained via reinforcement learning to generate an optimal target architecture. The weights of the target network are obtained by standard training, not by generation.

Neural networks that generate their own weights have been discussed in \citep{schmidhuber1993self,chang2018neural}. These may have interesting implications, such as the ability of a netowrk to self-introspect or self-replicate.

There are also Bayesian approaches for weight generation, which do not use a neural network as a generator. Markov Chain Monte Carlo \citep{neal1995bayesian,welling2011bayesian} use a Markov chain in the weight space with equilibrium distribution that equals the required posterior. In \citep{graves2011practical,blundell2015weight}, the posterior is approximated as a diagonal Gaussian with trainable means and variances, and the VI \citep{hinton1993keeping,kingma2017variational} objective is used. Another VI method is \citep{gal2016dropout}, which approximates the posterior of the weights as proportional to Bernoulli variables, which is equivalent to the dropout regularization technique \citep{srivastava2014dropout}.

\clearpage
\section{Another Way to Obtain the Hypernetwork Loss Function}\label{sec:relaxation}
Here we demonstrate how the loss function (\ref{eq:l}) arises from another consideration. We denote by $p_\varphi\left( \theta\right)$ the probability distribution over $\Theta$ of $\theta=G\left(z;\varphi\right)$. We also denote by $p\left(\theta|p_\text{data}\right)$ the required distribution over $\Theta$, which we would like $p_\varphi\left( \theta\right)$ to be equal to. An optimistic choice for $p\left( \theta|p_\text{data}\right)$ would be the following:
\begin{equation} \label{eq:pp}
p\left(\theta|p_\text{data}\right) =
\begin{cases}
\frac{1}{Z}       & \quad \text{if } \mathcal{L}\left(\theta|p_\text{data}\right) = \min_{\theta^\prime}\mathcal{L}\left(\theta^\prime|p_\text{data}\right)\\
0  & \quad \text{otherwise}
\end{cases}
\end{equation}
where $Z$ is a normalization constant. In other words, this distribution chooses only from the global optima, with equal probabilities. We can relax this optimistic form by turning it into a Gibbs distribution:
\begin{equation} \label{eq:p}
p\left(\theta|p_\text{data}\right)=\frac{\exp(-\lambda\mathcal{L}\left(\theta|p_\text{data}\right))}{Z},
\end{equation}
The hyperparameter $\lambda>0$ controls how close this distribution is to the optimistic form, which is recovered for $\lambda \rightarrow \infty$. The relaxation is required so that there is higher connectivity between the mode regions of $p\left(\theta|p_\text{data}\right)$ . This will make it possible for $p_\varphi\left( \theta\right)$ to become close to $p\left(\theta|p_\text{data}\right)$. Achieving this can be done by minimizing a loss function which is the Kullback-Leibler divergence between them:
\begin{equation} \label{eq:kl}
L=D_\text{KL}(p_\varphi\left( \theta\right)\|p(\theta|p_\text{data}))=\int p_\varphi\left( \theta\right) \log \left( \frac{p_\varphi(\theta)}{p(\theta|p_\text{data})}\right) \text{d}\theta
\end{equation}
(This integral can be defined also when $p_\varphi(\theta)$ is supported on a low dimensional manifold. In this case, $p_\varphi(\theta)$ can be written as a product of a delta function distribution and a finite distribution, where the delta serves as the restriction to the low dimensional manifold. The integration is understood to be only over this manifold, using the finite component of $p_\varphi(\theta)$. It can be seen from (\ref{eq:pp}) that $p(\theta|p_\text{data})$ never vanishes on this manifold). Inserting (\ref{eq:p}) into (\ref{eq:kl}) we get:
\begin{equation}\label{eq:lll}
L= \lambda \mathbb{E}_{\theta \sim p_\varphi} \mathcal{L}\left( \theta |p_\text{data}\right)+\mathbb{E}_{\theta \sim p_\varphi} \log p_\varphi(\theta)  + \log Z .
\end{equation}
Noticing that $\log Z$ is a constant (does not depend on $p_\varphi$), we see that (\ref{eq:lll}) is of the form of (\ref{eq:l}), where the diversity term is taken as the negation of the entropy.

\clearpage
\section{Relation to Variational Inference}\label{sec:vi}
The objective function for hypernetworks using VI \citep{krueger2017bayesian,louizos2017multiplicative} is:
\begin{eqnarray} \label{eq:l_vi1}
\tilde{L}_\text{VI}\left( \varphi\vert p_\text{noise},D\right) = -\mathbb{E}_{z \sim p_\text{noise}} [\log p(D\vert G\left(z;\varphi\right))\nonumber\\
-\log p_\text{prior}(G\left(z;\varphi\right)) + \log p(G\left(z;\varphi\right))],
\end{eqnarray}
where $D=\{(x_1,y_1),...,(x_n,y_n)\}$ is a data set of size $n$, $p_\text{prior}$ is a prior. The last term on the right hand side of equation (\ref{eq:l_vi1}) is the probability distribution induced by the hypernetwork. The first term is the probability computed by the target network. Assuming iid samples and conditional independence, we have:
\begin{eqnarray} \label{eq:iid}
\log p(D\vert G\left(z;\varphi\right))  &\equiv&  \log p(y_1,...,y_n\vert x_1,...,x_n;G\left(z;\varphi\right))\nonumber\\
&=& \sum_{i=1}^{n}\log p(y_i\vert x_i;G\left(z;\varphi\right)) .
\end{eqnarray}
We see that $\log p(D\vert G\left(z;\varphi\right))$ is an unbiased estimate of $n\mathbb{E}_{(x,y)\sim p_{\text{data}}}\log p(y\vert x;G\left(z;\varphi\right))$. Writing the objective in terms of ``true'' (as opposed to estimated) terms, we get:
\begin{eqnarray} \label{eq:l_vi2}
L_\text{VI}\left( \varphi\vert p_\text{noise},D\right) = -\mathbb{E}_{z \sim p_\text{noise},(x,y)\sim p_{\text{data}}}
[n\log p(y\vert x; G\left(z;\varphi\right)) \nonumber\\
-\log p_\text{prior}(G\left(z;\varphi\right)) + \log p(G\left(z;\varphi\right))],
\end{eqnarray}
The last term in this equation is the differential entropy of the generated weights. We see that equation (\ref{eq:l_vi2}) differs from equation (\ref{eq:ll}) only in the presence of the prior term. However, we note that differential entropy is not a correct generalization of Shannon entropy, and it has some undesired properties\citep{marsh2013introduction}. Generalizing from the discrete case to the continuous case requires either binning of the sample space, or using a reference probability distribution. The later approach yields the relative entropy (which is identical to the Kullback-Leibler divergence). Therefore, equation (\ref{eq:l_vi2}) can be seen as equivalent to equation (\ref{eq:ll}) if the entropy $\mathbb{H}$ in equation (\ref{eq:ll}) is measured with respect to a reference distribution $p_\text{prior}$.

\clearpage
\section{The Hyperparameter $\lambda$ in Variational Inference}\label{sec:vi2}
In this section, we explain why the hyperparameter $\lambda$ that appears in equation \ref{eq:l} could have also appeared in the objective for hypernetworks derived from VI\citep{krueger2017bayesian,louizos2017multiplicative}.

We look first at the non-Bayesian case. Denote by $p_\theta (y \vert x)$ the probability distribution computed by the target network. $p_\theta (y \vert x)$ is a function approximator, with weights $\theta$. In other words, $p_\theta (y \vert x)$ represents a family of models parametrized by $\theta$. A main idea in machine learning is to use a flexible function approximator, such that some weights $\theta^*$ yield good approximations to the ``true'' probability distribution (This is in contrast with other fields, such as physics, where a mininal model is preferred, derived from an underlying theory and assumptions). For predictive applications, it does not matter what the ontological interpretation of the model $p_{\theta^*} (y \vert x)$ is, so long as it gives good approximations.

In the Bayesian case, the weights are considered random variables, with a prior distribution $p(\theta)$. To emphasize this, we promote the subscript $\theta$ in $p_\theta (y \vert x)$ to the argument of the target network: $p(y \vert x;\theta)$. Notice that this is not a function approximator anymore - it is one single specific model that describes how $\theta$ and $x$ are combined to form the distribution over $y$ (The distributions $p(\theta)$ and $p(y \vert x;\theta)$ jointly form a generative model). There are no parameters to optimize so as to get a ``better'' model. Moreover, this model is observed only for one (unknown) sample of $\theta$. What is the ontological status of this model? If we treat this model as ``true'', then we can infer the posterior distribution using Bayes' law: $\log p(\theta\vert D)=\log p(D\vert \theta)+\log p(\theta)-\log Z$ for a normalizing factor $Z$ ($D$ is the data set, see Appendix \ref{sec:vi}). But usually the model $p(y \vert x;\theta)$ was not chosen on the basis of a theory or assumptions about the data, and for almost all values of $\theta$ it is meaningless to speak of the validity of $p(y \vert x;\theta)$ as a model. It follows that we should treat the model as a mere approximation. When inferring the posterior, we should take into consideration the amount of trust that we have in the model. One way of doing this is by adding the hyperparameter $\lambda$ to Bayes' law $\log p(\theta|D)=\lambda\log p(D|\theta)+\log p(\theta)-\log Z$, and the hyperparameter $\lambda$ would also appear in the VI objective.

\clearpage
\section{Entropy Estimation}\label{sec:entropy}
To backpropagate through the entropy term, using minibatches of samples of $z$, we will need to use an estimator for differential entropy. We use the Kozachenko-Leonenko estimator\citep{kozachenko1987sample,kraskov2004estimating}. For a set of samples $\theta_1,\theta_2,...,\theta_N$, the estimator is:
\begin{equation}
\hat{H}=\psi(N)+\frac{d}{N}\sum_{i=1}^{N}\log(\epsilon_i) ,
\end{equation}
where $\psi$ is the digamma function, $d$ is the dimension of the samples, which we take to be the dimension of $z$, and $\epsilon_i$ is the distance from $\theta_i$ to its nearest neighbor in the set of samples. The original estimator takes $d$ to be the dimension of $\theta$. However, as explained in Appendix \ref{sec:relaxation}, we are interested in an estimation of the differential entropy only on the manifold defined by $G(z)$, stripping away the delta functions that restrict $\theta$ to the manifold. We are using the Euclidean metric on the space $\Theta$, although a better estimator would use the metric induced on the lower dimensional manifold. We don't do this due to the complications that it introduces. We omit terms in the definition of the estimator which are constants that do not matter for optimization (The reason that we did keep the terms $\psi(N)$ and $\frac{d}{N}$ is that $N$ represents the batch size during training, but it also represents the size of the validation set during validation. If we want the entropy units to be comparable between the two, we need to keep the dependence on $N$, even though it does not matter for the optimization). This estimator is biased~\citep{charzynska2015improvement} but consistent in the mean square. In the case of the LeNet and ResNet-32 trials, we estimate the entropy term based on distances between on a subset of random samples for computational efficiency (and do not evaluate the full Cartesian product of pair-wise distances).

\clearpage
\section{Target Network Architectures}\label{sec:mnist_target}

\begin{table}[h]
	\caption{The target network architecture for MNIST. Note that all layers also have bias parameters. See \citep{krizhevsky2012imagenet,simonyan2014very} for definitions of the various terms.}
	{\footnotesize
		\begin{center}
			\begin{tabular}{| m{0.14\linewidth} | b{0.35\linewidth} | b{0.12\linewidth} | p{0.18\linewidth} |}
				\hline
				\textbf{Layer name} & \textbf{Layer components} & \textbf{Number of weights} & \textbf{Output size} \\ \hline
				input image &   &   & $28\times28\times1$ \\ \hline
				\multirow{3}{*}{layer1}
				& convolution: $5\times 5$,  stride: $1\times 1$,  padding: ‘SAME’,  number of filters: $32$  & 832 &  \\ \cline{2-2}
				& activation: ReLU & &  \\ \cline{2-2}
				& max pool: $2\times 2$,  stride: $2\times 2$,  padding: ‘SAME’ & & $14\times 14\times 32$ \\  \hline
				\multirow{3}{*}{layer2}
				& convolution: $5\times 5$,  stride: $1\times 1$,  padding: ‘SAME’,  number of filters: $16$  & 12,816 &  \\ \cline{2-2}
				& activation: ReLU & &  \\ \cline{2-2}
				& max pool: $2\times 2$,  stride: $2\times 2$,  padding: ‘SAME’ &  & $7\times 7\times 16$ \\ \hline
				\multirow{2}{*}{layer3}
				& fully-connected, number of filters: $8$  & 6,280 &  \\ \cline{2-2}
				& activation: ReLU &  & $8$ \\ \hline
				\multirow{2}{*}{layer4}
				& fully-connected, number of filters: $10$  & 90 &  \\ \cline{2-2}
				& softmax & & $10$ \\
				\hline
			\end{tabular}
		\end{center}
		\label{tab:target}
	}
\end{table}

\begin{table}[t]
	\caption{LeNet}
	{\footnotesize
		\begin{center}
			\begin{tabular}{| m{0.14\linewidth} | b{0.35\linewidth} | b{0.12\linewidth} | p{0.18\linewidth} |}
				\hline
				\textbf{Layer name} & \textbf{Layer components} & \textbf{Number of weights} & \textbf{Output size} \\ \hline
				input image &   &   & $32\times32\times3$ \\ \hline
				\multirow{3}{*}{layer1}
				& convolution: $5\times 5$,  stride: $1\times 1$,  padding: ‘SAME’,  number of filters: $6$  & 456 &  \\ \cline{2-2}
				& activation: ReLU & &  \\ \cline{2-2}
				& max pool: $2\times 2$,  stride: $2\times 2$& & $14\times 14\times 6$ \\  \hline
				\multirow{3}{*}{layer2}
				& convolution: $5\times 5$,  stride: $1\times 1$,  padding: ‘SAME’,  number of filters: $16$  & 2,416 &  \\ \cline{2-2}
				& activation: ReLU & &  \\ \cline{2-2}
				& max pool: $2\times 2$,  stride: $2\times 2$&  & $5\times 5\times 16$ \\ \hline
				\multirow{2}{*}{layer3}
				& fully-connected, number of filters: $120$  & 48,120 &  \\ \cline{2-2}
				& activation: ReLU &  & $120$ \\ \hline
				\multirow{2}{*}{layer4}
				& fully-connected, number of filters: $84$  & 10,164 &  \\ \cline{2-2}
				& activation: ReLU &  & $84$ \\ \hline
				\multirow{2}{*}{layer5}
				& fully-connected, number of filters: $10$  & 850 &  \\ \cline{2-2}
				& softmax & & $10$ \\
				\hline
			\end{tabular}
		\end{center}
		\label{tab:target}
	}
\end{table}

\begin{table}[h]
	\caption{ResNet-32}
	{\footnotesize
		\begin{center}
			\begin{tabular}{| m{0.14\linewidth} | b{0.25\linewidth} | b{0.22\linewidth} | p{0.18\linewidth} |}
				\hline
				\textbf{Layer name} & \textbf{Layer components} & \textbf{Number of weights} & \textbf{Output size} \\ \hline
				input image &   &   & $32\times32\times3$ \\ \hline
				\multirow{3}{*}{layer1}
				& convolution: $3\times 3$,  stride: $1\times 1$,  padding: ‘SAME’,  number of filters: $16$  & 448 &  \\ \cline{2-2}
				& activation: ReLU & & $32\times 32\times 16$ \\  \hline
				\multirow{3}{*}{layer2}
				& convolution: $3\times 3$,  stride: $1\times 1$,  padding: ‘SAME’,  number of filters: $16$  & $2,320\times 10 =  23,200$&  \\ \cline{2-2}
				& activation: ReLU  &  & $32\times 32\times 16$ \\ \hline
				\multirow{3}{*}{layer3}
				& convolution: number of filters: $32$  & $4,640 + 9,248 + 544 + 9,248\times 8 = 88,416$  &  \\ \cline{2-2}
				& activation: ReLU  &  & $16\times 16\times 32$ \\ \hline
				\multirow{3}{*}{layer4}
				& convolution: number of filters: $64$  & $18,496 + 36,928 + 2112 + 36,928\times 8 = 352,960$&  \\ \cline{2-2}
				& activation: ReLU & &  \\ \cline{2-2}
				& avg pool: $8\times 8$,  stride: $1\times 1$,  padding: $0$ &  & $1\times 1\times 64$ \\ \hline				  
				\multirow{2}{*}{layer5}
				& fully-connected, number of filters: $10$  & 650 &  \\ \cline{2-2}
				& softmax & & $10$ \\
				\hline
			\end{tabular}
		\end{center}
		\label{tab:target}
	}
\end{table}

\clearpage
\section{MNIST Hypernetwork Architecture and Training}\label{sec:mnist_hyper}
The specifications of the architecture for the hypernetwork are given in Table \ref{tab:blocks}. We choose the input $z$ to the network to be a $300$ dimensional vector, drawn from a uniform distribution over $\left[-1,1\right]^{300}$. The code sizes for the weight generators are chosen to be $15$ for all four layers. The extractor and the weight generators are all MLPs with leaky-ReLU activations~\citep{maas2013rectifier}. Batch normalization~\citep{ioffe2015batch} is employed in all layers besides the output layers of each sub-network. We do not use bias parameters in the hypernetwork, since our empirical evidence suggests (albeit inconclusively) that this helps for diversity. The total number of parameters in $G$ is $633,640$.

We trained the hypernetwork using the Adam optimizer~\citep{kingma2014adam}. The gradients of the loss were estimated using minibatches of $32$ samples of $z$, and $32$ images per sample of $z$ (we use different images for each noise sample). We found that this relatively large batch size is a good operating point in terms of the tradeoff between estimator variance and learning rate. We trained on a total of $13,000$ minibatches, and this took about $30$ minutes on a single NVIDIA Tesla K80 GPU.

\begin{table}[h]
	\caption{The layer structures for each of the fully-connected sub-networks of the hypernetwork. The layer structures are in the format $\text{(input size)}\rightarrow \text{(first layer size)} \rightarrow \text{(second layer size)} \rightarrow \text{(output size)} $. The sub-networks do not include bias terms. The number of parameters shown here does not include the batch normalization parameters.}
	{\footnotesize
		\begin{center}
			\begin{tabular}{| c | c | c |}
				\hline
				\textbf{Sub-net} & \textbf{Layer structure} & \textbf{Parameters} \\ \hline
				$E$ & \makecell{$300 \rightarrow 300 \rightarrow 300 \rightarrow  855$\\$=15\cdot(32+16+8+1)  $} & 436,500 \\ \hline
				$W_1$ & \makecell{$15 \rightarrow 40 \rightarrow 40\rightarrow 26 $\\$=5\cdot5+1$}  & 3240 \\ \hline
				$W_2$ & \makecell{$15 \rightarrow 100 \rightarrow 100 \rightarrow 801$\\$=5\cdot5\cdot32+1$}  & 91,600 \\ \hline
				$W_3$ & \makecell{$15 \rightarrow 100 \rightarrow 100\rightarrow 785$\\$=\left((28^2)/(4^2)\right)\cdot16+1$}  & 90,000 \\ \hline
				$W_4$ & \makecell{$15 \rightarrow 60 \rightarrow 60 \rightarrow 90$\\$=(8+1)\cdot10$}  & 9,900 \\
				\hline
			\end{tabular}
		\end{center}
		\label{tab:blocks}
	}
\end{table}

\clearpage
\section{CIFAR Hypernetwork Architecture and Training}\label{sec:cifar_hyper}
The specifications of the architecture for the hypernetwork are given in Table \ref{tab:blocks_cifar_lenet} and  \ref{tab:blocks_cifar_resnet} for the LeNet and ResNet-32 target network, respectively. We choose the input $z$ to the network to be a $100$ and $500$ dimensional vector for LeNet and ResNet-32, respectively, drawn from a standard normal distribution. The code sizes are determined by multiplying the number of filters in a given layer by a constant (here $8$). 

We trained the hypernetwork using the Adam optimizer~\citep{kingma2014adam} with a learning rate of $1\text{e-}4$ and $\beta=(0.9, 0.999)$. The gradients of the loss were estimated using minibatches of $10$ samples of $z$, and $512$ images per sample of $z$ (we use different images for each noise sample). 

For ResNet-32, we employed skip connections in the extractor and weight generators to ease optimization. Further, the weight generators output scale $\gamma$ and shift $\beta$ parameters of the batch-normalization operations~\citep{ioffe2015batch}. In order to apply batch-normalization in inference mode, we first compute moving averages of mean and variance statistics with one additional full pass over the training data.

\begin{table}[h]
	\caption{LeNet}
	{\footnotesize
		\begin{center}
			\begin{tabular}{| l | l | l |}
				\hline
				\textbf{Sub-net} & \textbf{Layer structure} & \textbf{Parameters} \\ \hline
				$E$ & \makecell{$100 \rightarrow 200 \rightarrow 200 \rightarrow  236$\\$=8\cdot(6+16+120+84+10)  $} & 107,200 \\ \hline
				$W_1$ & \makecell{$8 \rightarrow 400 \rightarrow 400\rightarrow 76 $\\$=5\cdot5\cdot3+1$}  & 193,600 \\ \hline
				$W_2$ & \makecell{$8 \rightarrow 400 \rightarrow 400 \rightarrow 151$\\$=5\cdot5\cdot6+1$}  & 223,600 \\ \hline
				$W_3$ & \makecell{$8 \rightarrow 400 \rightarrow 400\rightarrow 401$\\$=5\cdot5\cdot16+1$}  & 323,600 \\ \hline
				$W_4$ & \makecell{$8 \rightarrow 400 \rightarrow 400\rightarrow 10,164$\\$=(120+1)\cdot84$}  & 4,228,800 \\ \hline
				$W_5$ & \makecell{$8 \rightarrow 400 \rightarrow 400 \rightarrow 850$\\$=(84+1)\cdot10$}  & 503,200 \\
				\hline
			\end{tabular}
		\end{center}
		\label{tab:blocks_cifar_lenet}
	}
\end{table}

\begin{table}[h]
	\caption{ResNet-32}
	{\footnotesize
		\begin{center}
			\begin{tabular}{| l | l | l |}
				\hline
				\textbf{Sub-net} & \textbf{Layer structure} & \textbf{Parameters} \\ \hline
				$E$ & \makecell{$500 \rightarrow 1000 \rightarrow 1000 \rightarrow  1104$\\$=8\cdot(16+16+32+64+10)  $} & 2,935,200 \\ \hline
				$W_1$ & \makecell{$8 \rightarrow 800 \rightarrow 800\rightarrow 10 $\\$=3\cdot3\cdot3+1$}  & 654,400 \\ \hline
				$W_2$ & \makecell{$8 \rightarrow 800 \rightarrow 800 \rightarrow 145$\\$=3\cdot3\cdot16+1$}  & 762,400 \\ \hline
				$W_3$ & \makecell{$8 \rightarrow 800 \rightarrow 800\rightarrow 145$\\$=3\cdot3\cdot16+1$}  & 762,400 \\ \hline
				$W_4$ & \makecell{$8 \rightarrow 800 \rightarrow 800\rightarrow 289$\\$=3\cdot3\cdot32+1$}  & 887,600 \\ \hline
				$W_5$ & \makecell{$8 \rightarrow 800 \rightarrow 800 \rightarrow 650$\\$=(64+1)\cdot10$}  & 1,166,400 \\
				\hline
			\end{tabular}
		\end{center}
		\label{tab:blocks_cifar_resnet}
	}
\end{table}

\clearpage
\section{Paths in Weight Space - MNFG}\label{sec:path_mnf}
\begin{figure}[h]
	\centering
	\begin{subfigure}{\linewidth}\centering\includegraphics[scale=1]{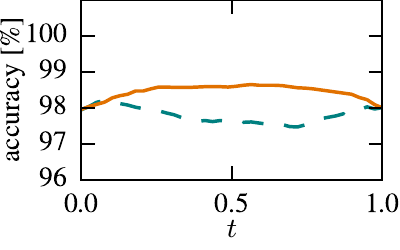}\end{subfigure}
	\begin{subfigure}{\linewidth}\centering\includegraphics[scale=1]{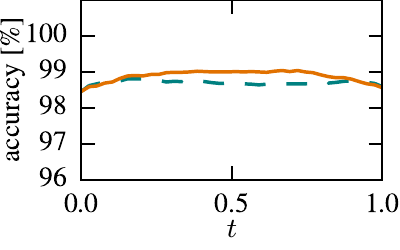}\end{subfigure}
	\begin{subfigure}{\linewidth}\centering\includegraphics[scale=1]{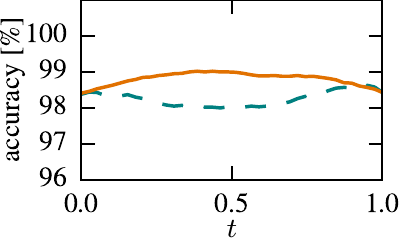}\end{subfigure}
	\caption{Accuracies along paths, with end points $z_1$, $z_2$ sampled at random, \textbf{for MNFG}. Each graph corresponds to a different sampled pair of endpoints. The dashed lines are the direct paths, and the solid lines are the interpolated paths.}
	\label{fig:paths_mnf}
\end{figure}

\clearpage
\section{PCA Scatter Plots}\label{sec:pca}
\begin{figure}[H]
	\centering
	\includegraphics[width=\linewidth]{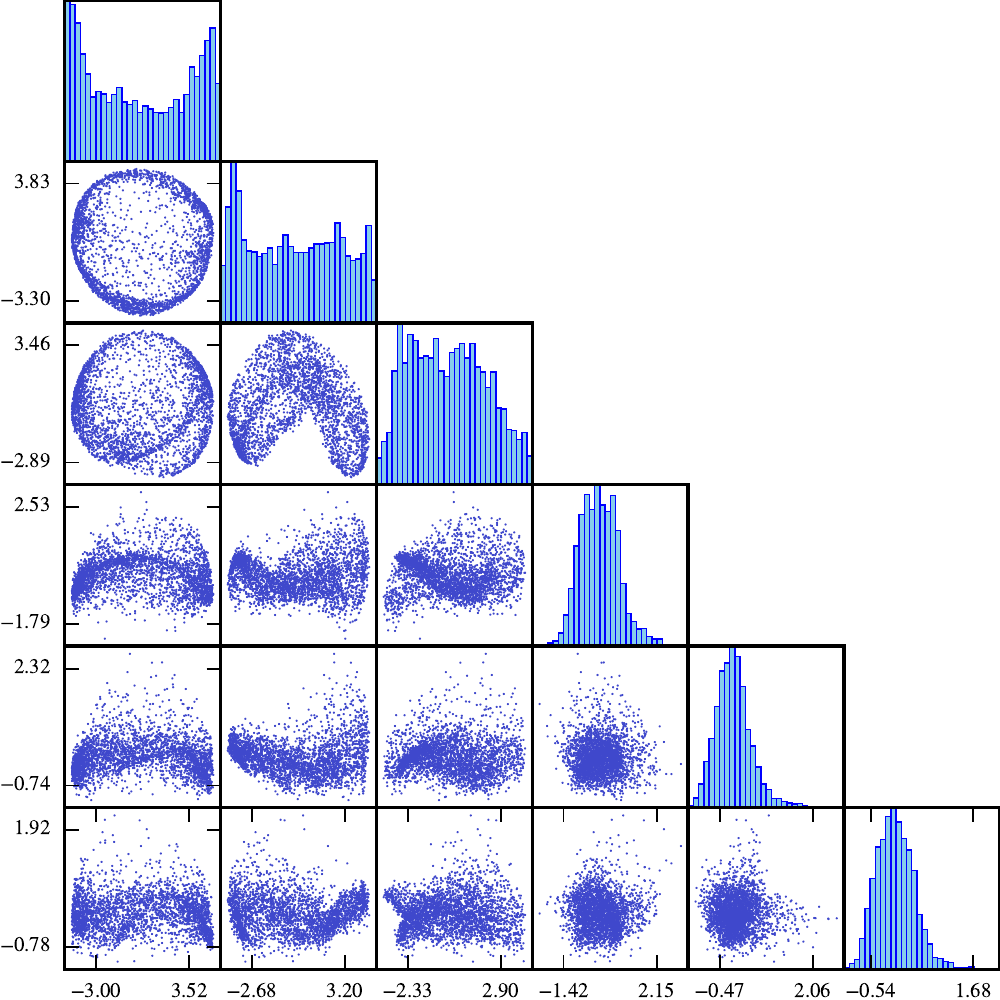}
	\caption{Scatter plots of the generated weights for a specific first layer filter, in PCA space. The $(i,j)$ scatter plot has principal component $i$ against principal component $j$.}
	\label{fig:mnist_layer1_filt_pca}
\end{figure}

\begin{figure}[H]
	\centering
	\includegraphics[width=\linewidth]{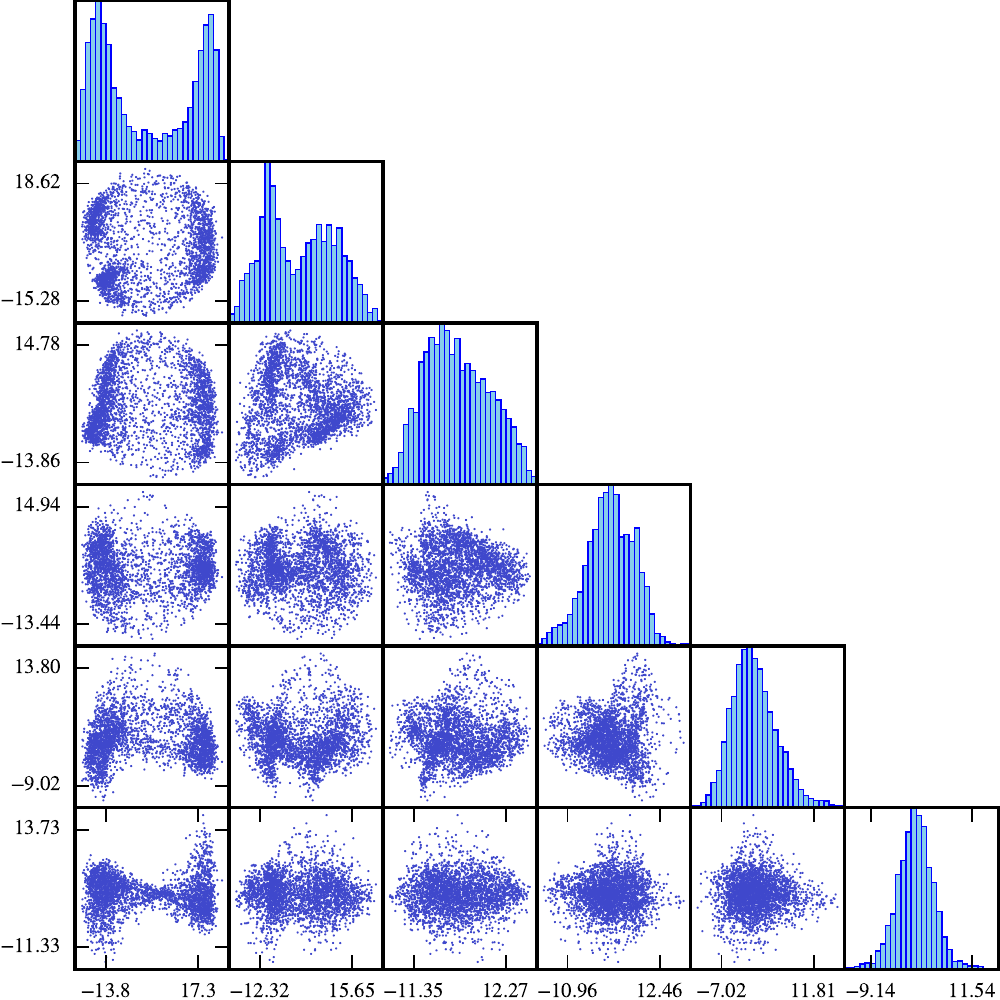}
	\caption{Scatter plots of the generated weights for a specific third layer filter, in PCA space. The $(i,j)$ scatter plot has principal component $i$ against principal component $j$.}
	\label{fig:mnist_layer3_filt_pca}
\end{figure}

\section*{}

\begin{figure}[H]
	\centering
	\includegraphics[width=\linewidth]{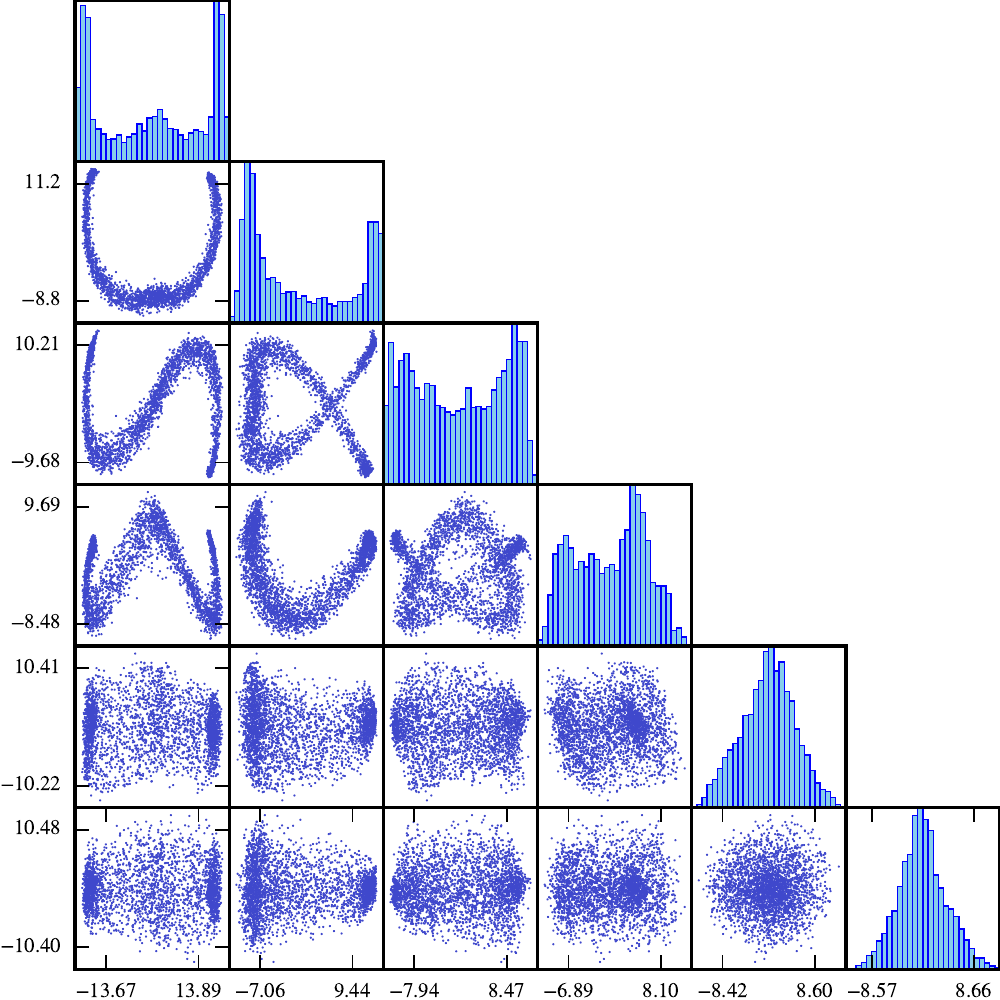}
	\caption{Scatter plots of the generated weights for the entire first layer, in PCA space. The $(i,j)$ scatter plot has principal component $i$ against principal component $j$.}
	\label{fig:mnist_layer1_pca}
\end{figure}

\begin{figure}[H]
	\centering
	\includegraphics[width=\linewidth]{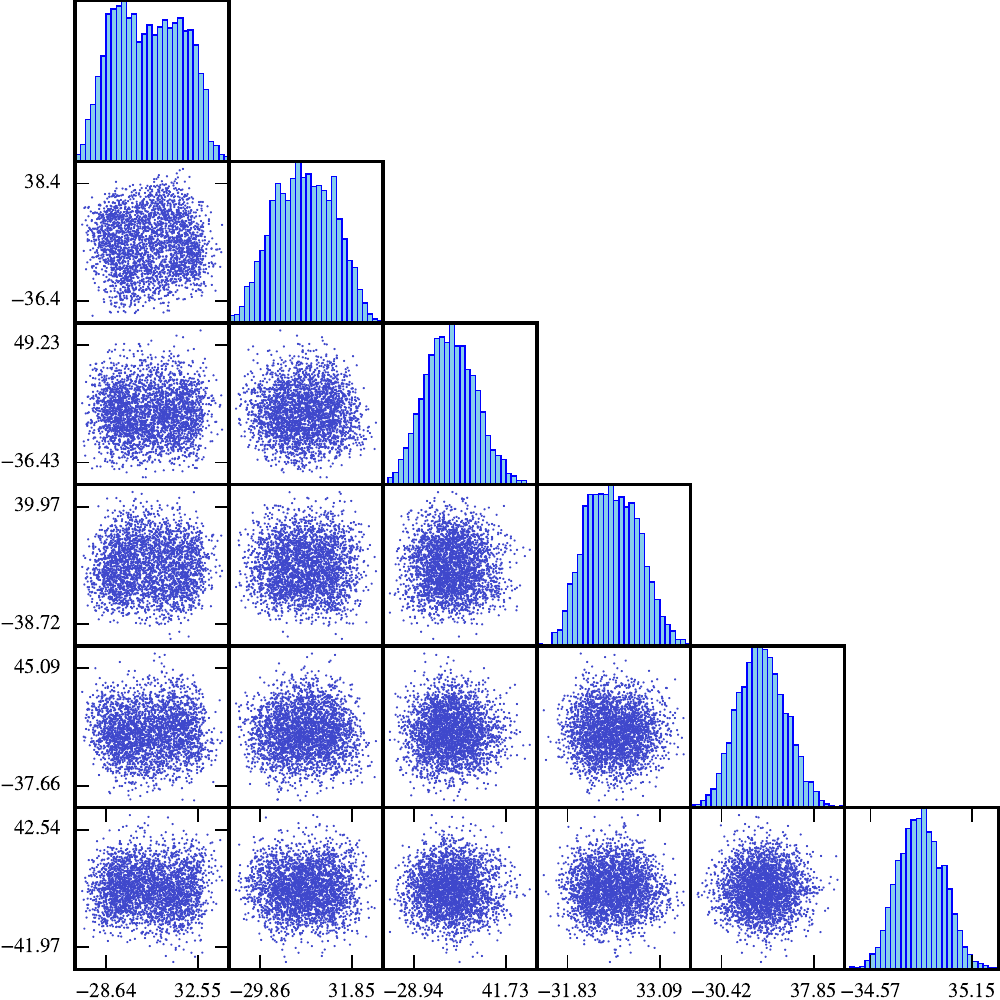}
	\caption{Scatter plots of the generated weights for the entire second layer, in PCA space. The $(i,j)$ scatter plot has principal component $i$ against principal component $j$.}
	\label{fig:mnist_layer2_pca}
\end{figure}

\begin{figure}[H]
	\centering
	\includegraphics[width=\linewidth]{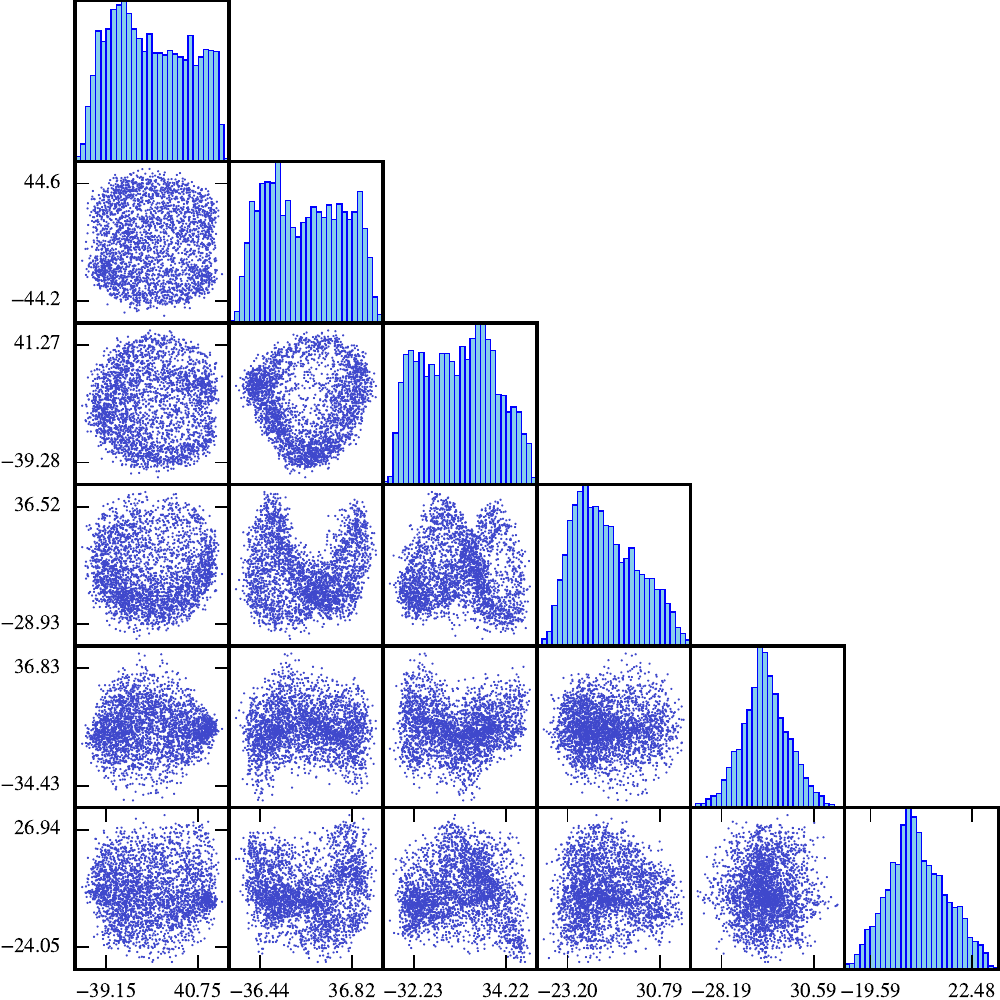}
	\caption{Scatter plots of the generated weights for the entire third layer, in PCA space. The $(i,j)$ scatter plot has principal component $i$ against principal component $j$.}
	\label{fig:mnist_layer3_pca}
\end{figure}

\begin{figure}[H]
	\centering
	\includegraphics[width=\linewidth]{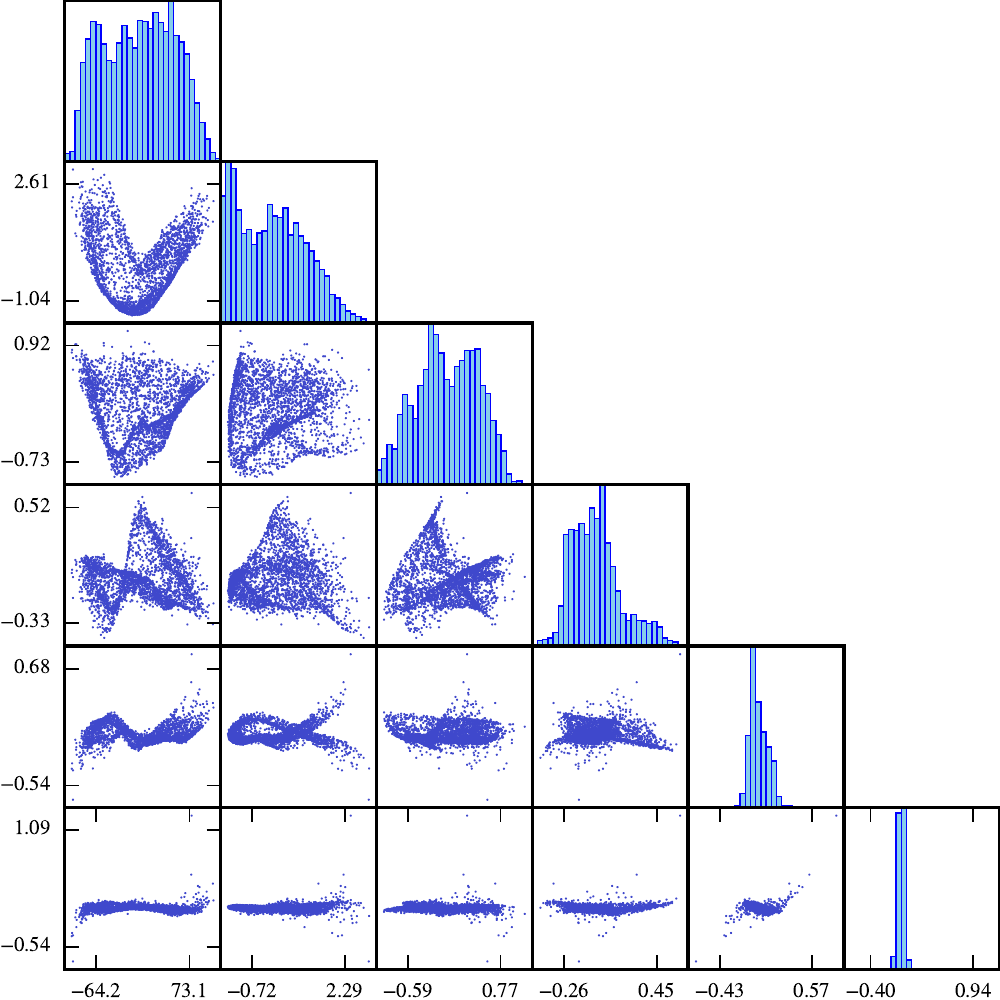}
	\caption{Scatter plots of the generated weights for the entire fourth layer, in PCA space. The $(i,j)$ scatter plot has principal component $i$ against principal component $j$.}
	\label{fig:mnist_layer4_pca}
\end{figure}

\begin{figure}[H]
	\centering
	\includegraphics[width=\linewidth]{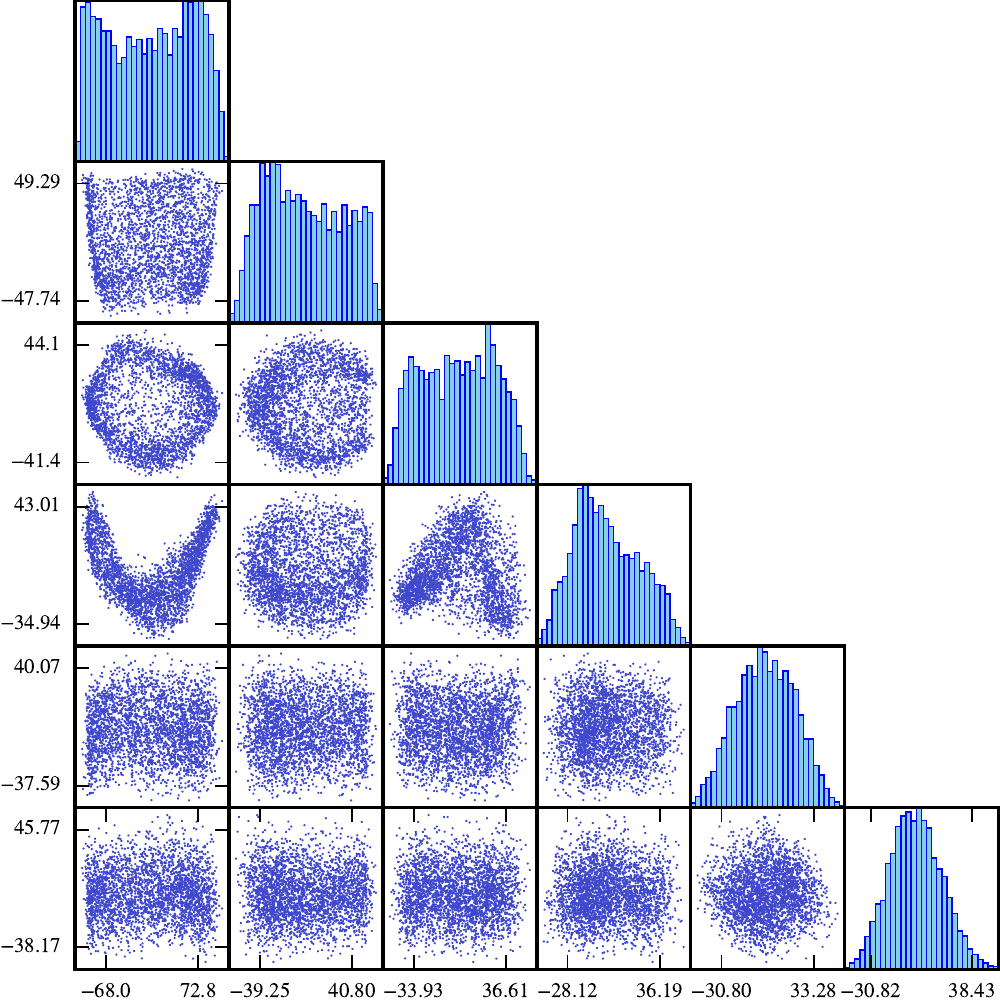}
	\caption{Scatter plots of the generated weights for the entire target network, in PCA space. The $(i,j)$ scatter plot has principal component $i$ against principal component $j$.}
	\label{fig:mnist_all_layers_pca}
\end{figure}

\clearpage
\section{PCA Scatter Plots - MNFG}\label{sec:pca_mnf}
\begin{figure}[H]
	\centering
	\includegraphics[width=\linewidth]{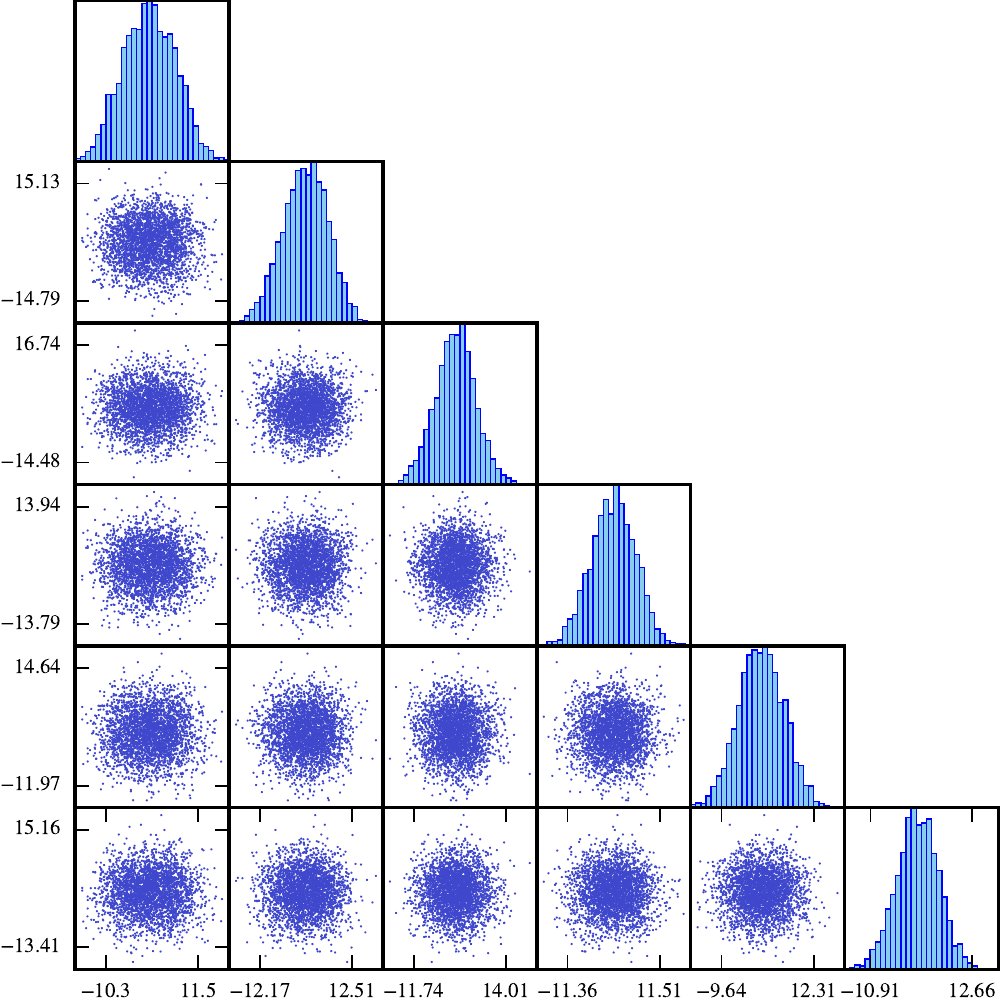}
	\caption{Scatter plots of the generated weights for the entire first layer, in PCA space, \textbf{for MNFG}. The $(i,j)$ scatter plot has principal component $i$ against principal component $j$.}
	\label{fig:mnist_layer1_pca_mnf}
\end{figure}

\begin{figure}[H]
	\centering
	\includegraphics[width=\linewidth]{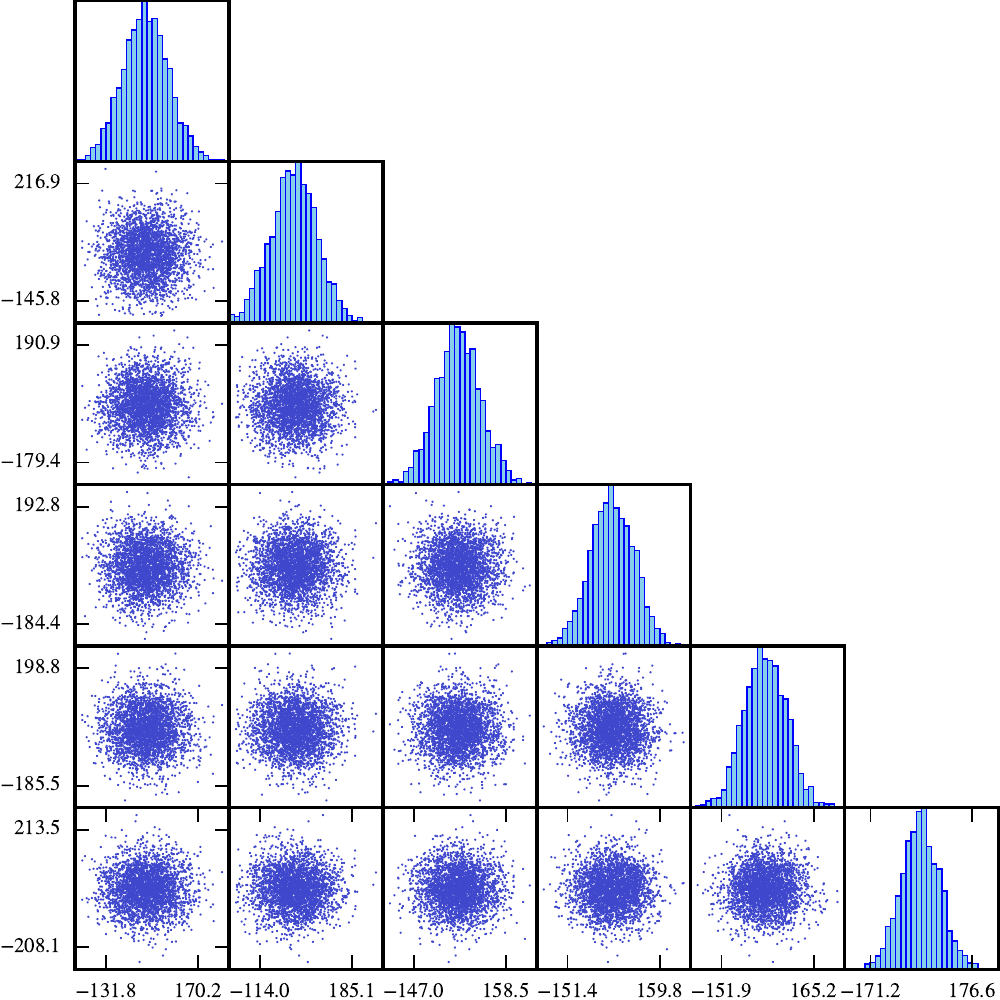}
	\caption{Scatter plots of the generated weights for the entire second layer, in PCA space, \textbf{for MNFG}. The $(i,j)$ scatter plot has principal component $i$ against principal component $j$.}
	\label{fig:mnist_layer2_pca_mnf}
\end{figure}

\begin{figure}[H]
	\centering
	\includegraphics[width=\linewidth]{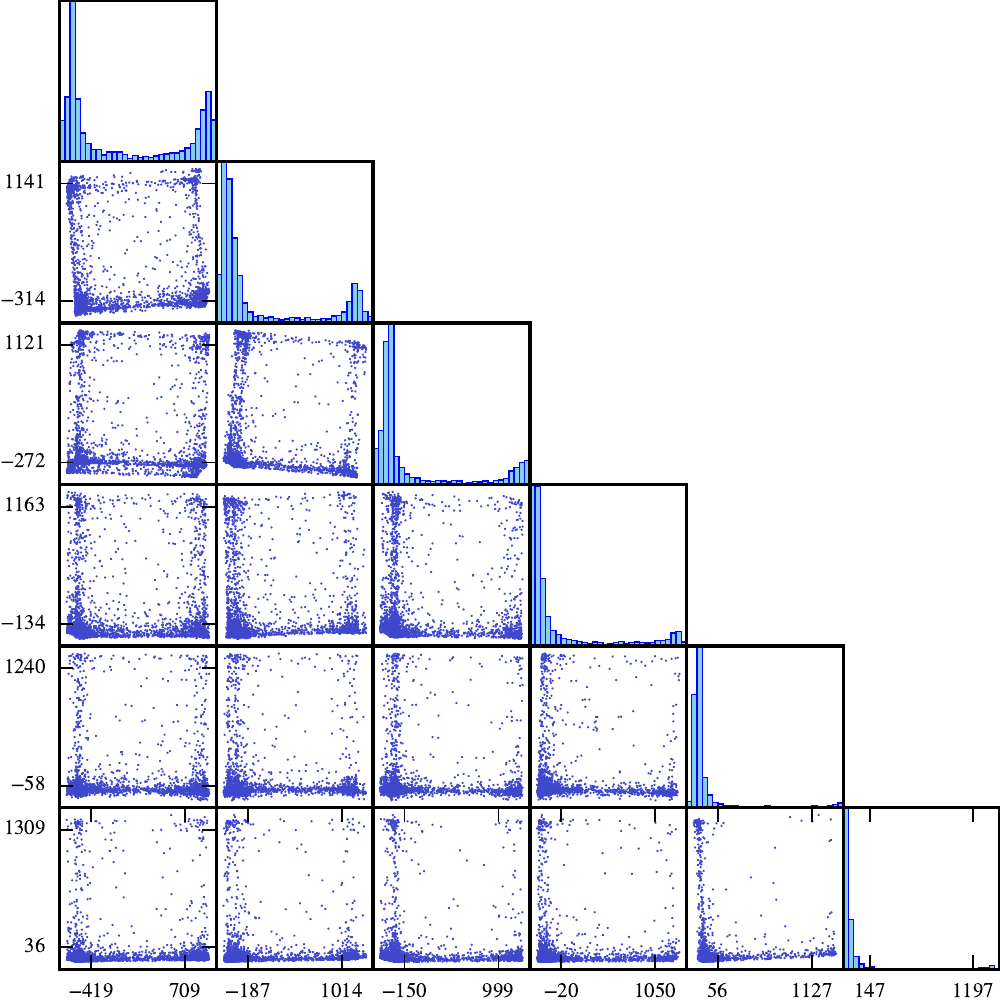}
	\caption{Scatter plots of the generated weights for the entire third layer, in PCA space, \textbf{for MNFG}. The $(i,j)$ scatter plot has principal component $i$ against principal component $j$.}
	\label{fig:mnist_layer3_pca_mnf}
\end{figure}

\begin{figure}[H]
	\centering
	\includegraphics[width=\linewidth]{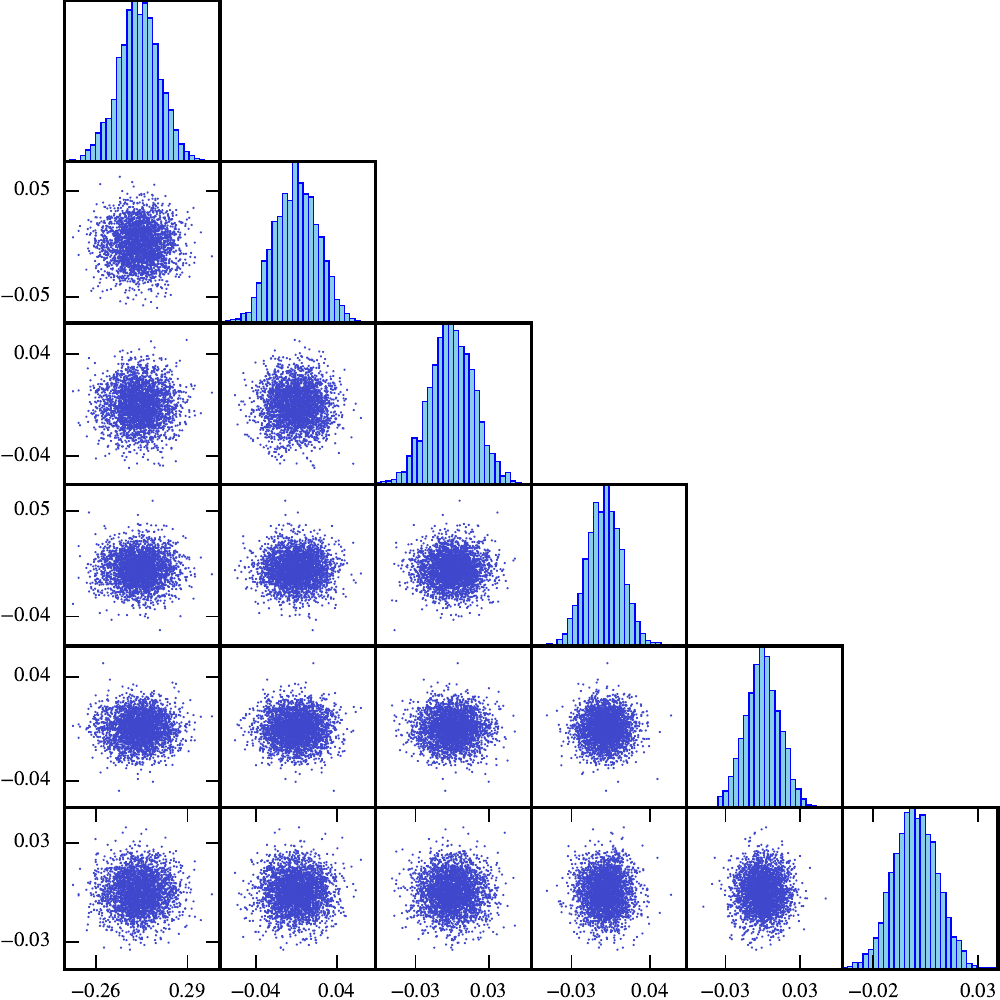}
	\caption{Scatter plots of the generated weights for the entire fourth layer, in PCA space, \textbf{for MNFG}. The $(i,j)$ scatter plot has principal component $i$ against principal component $j$.}
	\label{fig:mnist_layer4_pca_mnf}
\end{figure}

\begin{figure}[H]
	\centering
	\includegraphics[width=\linewidth]{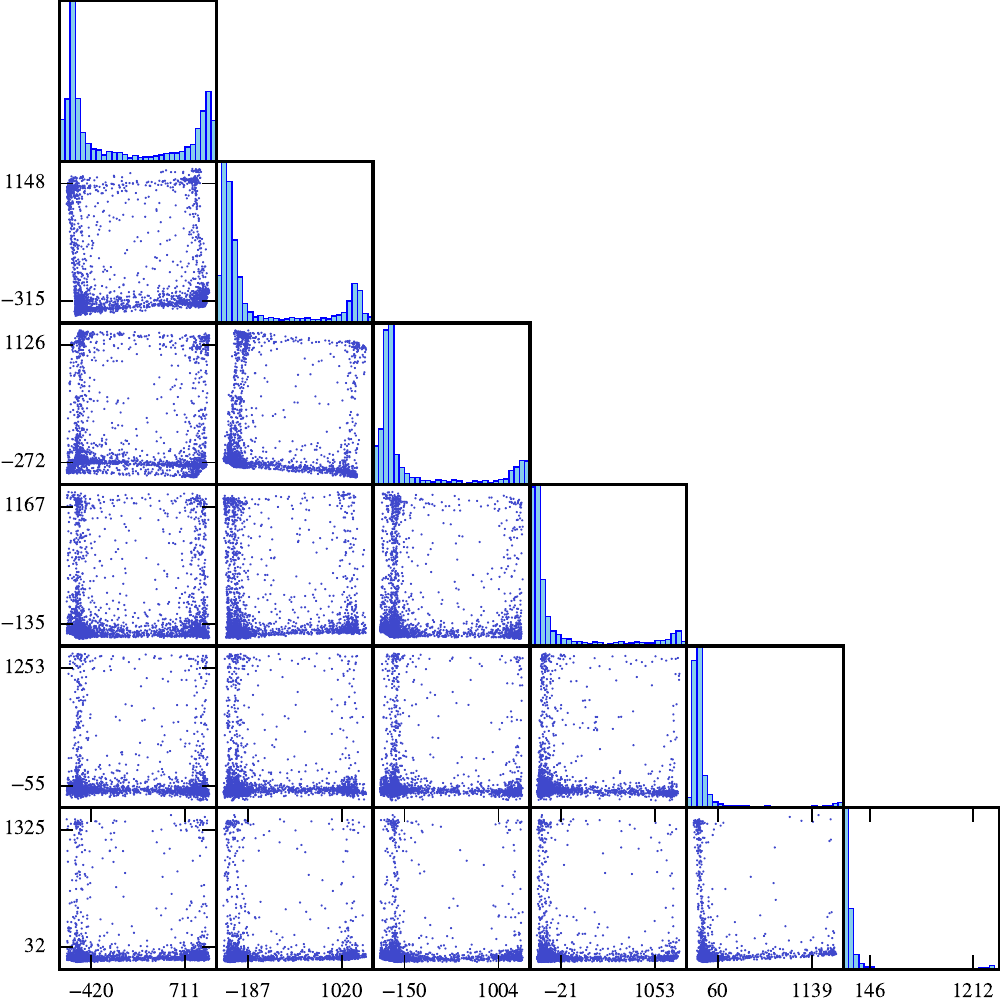}
	\caption{Scatter plots of the generated weights for the entire target network, in PCA space, \textbf{for MNFG}. The $(i,j)$ scatter plot has principal component $i$ against principal component $j$.}
	\label{fig:mnist_all_layers_pca_mnf}
\end{figure}

\end{document}